\documentclass{article}

\usepackage[final]{neurips_2025}

\usepackage[utf8]{inputenc} % allow utf-8 input
\usepackage[T1]{fontenc}    % use 8-bit T1 fonts
\usepackage{hyperref}       % hyperlinks
\usepackage{url}            % simple URL typesetting
\usepackage{booktabs}       % professional-quality tables
\usepackage{amsfonts}       % blackboard math symbols
\usepackage{nicefrac}       % compact symbols for 1/2, etc.
\usepackage{microtype}      % microtypography
\usepackage{xcolor}         % colors
\usepackage{multirow}
\usepackage{graphicx}
\usepackage{enumitem}
\usepackage{amsmath}
\usepackage{wrapfig}
\usepackage{rotating} 
\usepackage[linesnumbered,ruled,lined]{algorithm2e}
\usepackage[normalem]{ulem}

\useunder{\uline}{\ul}{}

\SetCommentSty{mycommfont}

\hypersetup{
    colorlinks=true,
    citecolor=teal,
    linkcolor=red,
    filecolor=magenta,
    urlcolor=magenta}

\newcommand{\mat}[1]{{\bf #1}}   % matrix: bold
\newcommand{\method}{\textsc{Dmmv}}
\newcommand{\methoda}{\textsc{Dmmv-s}}
\newcommand{\methodb}{\textsc{Dmmv-a}}
\newcommand{\maskmethod}{\textsc{BCmask}}
\newcommand{\red}[1]{\textcolor{red}{#1}}

\definecolor{yucolor}{HTML}{8A2BE2}

\title{Multi-Modal View Enhanced Large Vision Models for Long-Term Time Series Forecasting}

\author{ChengAo Shen\textsuperscript{\rm 1}, Wenchao Yu\textsuperscript{\rm 2}, Ziming Zhao\textsuperscript{\rm 1}, Dongjin Song\textsuperscript{\rm 3}, Wei Cheng\textsuperscript{\rm 2},\\
\textbf{Haifeng Chen\textsuperscript{\rm 2}, Jingchao Ni\textsuperscript{\rm 1}}\\
\textsuperscript{\rm 1}University of Houston, \textsuperscript{\rm 2}NEC Laboratories America, \textsuperscript{\rm 3}University of Connecticut\\
\textsuperscript{\rm 1}\texttt{\{cshen9,zzhao35,jni7\}@uh.edu}, \textsuperscript{\rm 2}\texttt{\{wyu,weicheng,haifeng\}@nec-labs.com},\\
\textsuperscript{\rm 3}\texttt{dongjin.song@uconn.edu}}

\begin{document}

\maketitle

\begin{abstract}

Time series, typically represented as numerical sequences, can also be transformed into images and texts, offering multi-modal views (MMVs) of the same underlying signal. These MMVs can reveal complementary patterns and enable the use of powerful pre-trained large models, such as large vision models (LVMs), for long-term time series forecasting (LTSF). However, as we identified in this work, the state-of-the-art (SOTA) LVM-based forecaster poses an inductive bias towards ``forecasting periods''. To harness this bias, we propose \method, a novel decomposition-based multi-modal view framework that leverages trend-seasonal decomposition and a novel backcast-residual based adaptive decomposition to integrate MMVs for LTSF. Comparative evaluations against 14 SOTA models across diverse datasets show that \method\ outperforms single-view and existing multi-modal baselines, achieving the best mean squared error (MSE) on 6 out of 8 benchmark datasets. The code for this paper is available at: \url{https://github.com/D2I-Group/dmmv}.
\end{abstract}

% \vspace{-0.3cm}

\section{Introduction}

\vspace{-0.05cm}

Long-term time series forecasting (LTSF) is vital across domains such as geoscience \cite{ardid2025ergodic}, neuroscience \cite{caro2024brainlm}, energy \cite{koprinska2018convolutional}, healthcare \cite{morid2023time}, and smart city \cite{ma2017learning}. %Recently, with the significant advances of sequence modeling in the language domain, growing research attentions on time series have been drawn to methods ranging from Transformer \cite{wen2023transformers} to Large Language Models (LLMs) \cite{jiang2024empowering,zhang2024large}.
Inspired by the success of Transformers and Large Language Models (LLMs) in the language domain, recent research has explored similar architectures for time series \cite{wen2023transformers,jiang2024empowering,zhang2024large}. %As
Meanwhile, Large Vision Models (LVMs) like \texttt{ViT} \cite{dosovitskiy2021image}, \texttt{BEiT} \cite{bao2022beit} and \texttt{MAE} \cite{he2022masked}, %become achieving a similar success as LLMs (but in vision domain), some of the emergent efforts have been invested to explore the potential of LVMs in time series forecasting \cite{chen2025visionts}.
have achieved comparable breakthroughs in the vision domain, prompting interest in their application to LTSF \cite{chen2025visionts}. %In these works, time series are imaged, {\em i.e.}, transformed to an image representation \cite{ni2025harnessing}, %as illustrated by Fig. \ref{fig.method}(a),
%then fed to an LVM to learn embeddings that can be decoded to forecasts.
These approaches transform time series into image-like representations, enabling LVMs to extract embeddings for forecasting \cite{ni2025harnessing}. %These works posit that LVMs, being pre-trained on vast images, can transfer useful knowledge to LTSF task because of the inherent relationship between images and time series -- each row/column in an image (per channel) is a sequence of {\em continuous} pixel values that resembles a univariate time series (UTS). This relationship makes LVMs more aligned with time series than LLMs since LLMs consume {\em discrete} tokens.
The rationale is that LVMs, pre-trained on large-scale image datasets, can transfer useful knowledge to LTSF due to a structural similarity: each image channel contains sequences of {\em continuous} pixel values analogous to univariate time series (UTS). This alignment suggests LVMs may be better suited to time series than LLMs, which process {\em discrete} tokens.

% This hypothesis is somewhat validated by the state-of-the-art (SOTA) \texttt{VisionTS} model \cite{chen2025visionts}, which successfully applies \texttt{MAE} \cite{he2022masked} on imaged time series and achieves remarkable forecasting performance. %competitive, sometimes superior, forecasting errors to other non-LVM models.
This hypothesis is partially validated by the SOTA \texttt{VisionTS} model \cite{chen2025visionts}, which applies \texttt{MAE} \cite{he2022masked} to imaged time series and achieves impressive forecasting performance. This progress has spurred interests in combining imaged time series with other representations.
%The astonishing progress propels the research in the direction of %adapting multi-modal models to
%integrating imaged time series with other modalities of time series.
In the past, %time series have been studied in different representations including
time series have been studied through various forms: %(1) numerical sequences (the original, raw time series) \cite{zeng2023transformers,nie2023time}, (2) imaged time series \cite{wu2023timesnet,chen2025visionts}, (3) verbalized time series ({\em i.e.}, texts) \cite{xue2023promptcast,gruver2023large}, {\em etc.}
(1) raw numerical sequences \cite{zeng2023transformers,nie2023time}, (2) imaged representations \cite{wu2023timesnet,chen2025visionts}, and (3) verbalized (textual) descriptions \cite{xue2023promptcast,gruver2023large}. %Despite their distinct modalities, these representations are basically different views of the same data. This distinguishes them from the so-called multimodal data, where different modalities are usually produced by different sources of the same system.
While they differ in modality, they represent alternative views of the same underlying data -- unlike typical multi-modal data, where modalities originate from distinct sources \cite{liu2023visual}. 
% However, the multi-modal nature of such views enables the use of large models, such as LLMs, LVMs, and vision-language models (VLMs) \cite{kim2021vilt,radford2021learning,liu2023visual}, that are pre-trained on vast data of other modalities for time series analysis.
However, these {\em multi-modal views (MMVs)} enable the application of large pre-trained models, such as LLMs, LVMs, and vision-language models (VLMs) \cite{kim2021vilt,radford2021learning}, to time series analysis, specializing them from those in conventional multi-view learning \cite{wang2015deep}, where multi-view is a broader notion including both MMVs and views of the same modality ({\em e.g.}, augmented image views \cite{wang2015deep}). To distinguish, we use MMVs for time series throughout this paper.
% This specializes such views from those in %traditional views in
% traditional multi-view learning \yu{not sure if we need to explicitly emphsize the difference since multi-view learning could have views from different modalities}, which include views of the same modality, {\em e.g.}, differently perturbed image views of the same object \cite{wang2015deep}. As such, in this work, we call such views as {\em multi-modal views} (MMV) of time series. \yu{Most classical multi‑view learning work confines itself to single‑modality scenarios -- that is, ....  Because the views we construct span distinct modalities -- ... -- we explicitly refer to them as *multi‑modal views (MMVs)* throughout this paper. (This way, we downplay the fact that we're creating a new term, so reviewers may judge it less harshly.)}

% The advantage of leveraging MMV is twofold. First, augmenting time series with views of other modalities could complement patterns that are not obvious in the (original) numerical view. Second, incorporating pre-trained large models could %facilitate
% enhance extraction of intricate patterns that a specific view ({\em e.g.}, visual view) harbor.
Leveraging MMVs offers two key advantages: (1) augmenting time series with alternative views can reveal patterns not evident in the original numerical data, and (2) pre-trained large models can extract complex patterns specific to certain views, such as visual representations. %In light of this, in this work, we are interested in investigating the synergy of multi-modal views in time series analysis. In particular, inspired by recent success of LVMs, we aim to explore the potential of LVMs in an MMV framework for LTSF task. To the best of our knowledge, integrating visual view of time series via LVMs with other views remains an underexplored problem.
Motivated by these benefits and the recent success of LVMs, this work investigates the synergy of MMVs for LTSF, with a focus on incorporating LVMs. To our knowledge, integrating the visual view of time series via LVMs alongside other modalities remains underexplored. %The most relevant is the recent \texttt{Time-VLM} model \cite{zhong2025time}, which employs a VLM {\em i.e.}, \texttt{ViLT} \cite{kim2021vilt}, as its backbone for encoding the visual view and some contextual texts, augmented with a transformer for encoding the numerical view. The embeddings of different views are fused into a single embedding by a fusion layer, which serves as the input to a prediction layer.
The most related effort, \texttt{Time-VLM} \cite{zhong2025time}, uses a VLM (\texttt{ViLT} \cite{kim2021vilt}) to encode visual view and contextual texts of time series, augmented with a \texttt{Transformer} for the numerical view. All embeddings are combined through a fusion layer. %However, this straightforward combination-based paradigm does not harness the inductive bias of individual views, leading to suboptimal results (as evaluated in $\S$\ref{sec.exp}).
However, this simple combination strategy overlooks the unique inductive biases of individual views, leading to suboptimal performance (see $\S$\ref{sec.exp.results}). %Also, as demonstrated by its ablation analysis, incorporating contextual texts brings marginal gains in performance, but could incur extra high computational costs due to the involvement of a large language encoder.
Moreover, its use of textual inputs provides only marginal improvements while introducing significant computational overhead due to the large language encoder.
% \yu{discuss VisionTS here as well, including contents from 'an inductive bias' section}

% In this work, we propose a \underline{D}ecomposition-based \underline{M}ulti-\underline{M}odal \underline{V}iew Framework (\method) for LTSF. \method\ is a compact model that integrates the numerical and visual views of time series.
We propose \method, a \underline{D}ecomposition-based \underline{M}ulti-\underline{M}odal \underline{V}iew Framework for LTSF, which integrates numerical and visual views in a compact architecture.
% Textual view is excluded not only because of the marginal gains of contextual cues as shown by \texttt{Time-VLM} \cite{zhong2025time}, but also for the recent argument about the effectiveness of LLMs in time series forecasting \cite{tan2024language}, particularly considering the high costs of LLMs.\yu{merge Time-VLM discussion into previous paragraph. this paragraph only discuss the proposed method?} %Considering the questionable utility of LLMs and their high fine-tuning costs, \method\ adopts a compact design.
We exclude the textual view due to its marginal gains in \texttt{Time-VLM} \cite{zhong2025time} and recent doubts about the effectiveness and cost-efficiency of LLMs for LTSF \cite{tan2024language}. % Basically, \method\ has compact design with a numerical forecaster and a visual forecaster. Inspired by \texttt{VisionTS} \cite{chen2025visionts}, the visual forecaster uses \texttt{MAE} \cite{he2022masked} -- a self-supervisedly pre-trained LVM that can reconstruct masked images -- as the backbone for its remarkable ability in predicting continuous values ({\em i.e.}, pixels). Time series are imaged using a period-based patching technique \cite{wu2023timesnet}, which, although shows effectiveness in \cite{chen2025visionts}, has an inductive bias towards forecasting time series with strong periodicity.
\method\ comprises two specialized forecasters: a numerical forecaster and a visual forecaster. The visual forecaster, inspired by \texttt{VisionTS} \cite{chen2025visionts}, uses \texttt{MAE} \cite{he2022masked} -- a self-supervised LVM capable of reconstructing masked images -- leveraging its strong performance on continuous values ({\em i.e.}, pixels). Time series are transformed into images using a period-based patching technique \cite{wu2023timesnet}, which, although effective, imposes an inductive bias on LVMs towards periodic signals. %As such, we design two \method\ instances to adapt the inductive bias \yu{better to refer and describe figure 1 here since the plot is not technical}: (1) \methoda\ (\underline{s}imple decomposition), which first decomposes an input time series into trend and seasonal components, then uses the numerical and visual forecasters to process the trend and seasonal components, respectively; (2) \methodb\ (\underline{a}daptive decomposition), which automatically learns trend and seasonal components using a backcast-residual mechanism to fit the two forecasters. %Unlike the {\em intermediate} fusion in \texttt{Time-VLM},
To address this, we design two \method\ variants as illustrated in Fig. \ref{fig.method}: (a) \methoda\ (\underline{s}imple decomposition), which splits the time series into trend and seasonal components, assigning them to the numerical and visual forecasters, respectively; (b) \methodb\ (\underline{a}daptive decomposition), which adaptively learns the decomposition via a backcast-residual mechanism aligned with the two forecasters. %\method\ adopts a {\em late} fusion strategy \cite{kline2022multimodal} to merge the two views by a gating mechanism due to the important role of the decoder of \texttt{MAE} in pixel prediction, which makes %embedding alone, and in turn intermediate fusion, suboptimal.
% {\em intermediate} fusion using embeddings suboptimal. Thorough experiments indicate \method\ substantially outperforms the SOTA single-view %uni-modal view
% baselines and \texttt{Time-VLM} although it has an additional text encoder.
\method\ employs {\em late fusion} \cite{kline2022multimodal} via a gating mechanism, as intermediate fusion ({\em e.g.}, embedding-level) underutilizes \texttt{MAE}'s decoder, which plays a crucial role in pixel prediction. Extensive experiments show that \method\ significantly outperforms both SOTA single-view methods and \texttt{Time-VLM}, despite the latter incorporating an additional text encoder. To sum up, our contributions are as follows.
\begin{itemize}[topsep=0pt,leftmargin=*]%[noitemsep,topsep=0pt,leftmargin=*]
\setlength{\itemsep}{0.01cm}
\item We distinguish MMVs in time series analysis from the broader notion in conventional multi-view learning and study the emergent yet underexplored problem of MMV-based LTSF.
% \item We propose \method, a novel MMV framework that is carefully designed to harness an inductive bias that we identified from the contemporary best LVM-based forecasters, and complement it with the strength of a numerical forecaster, instantiating two variants \methoda\ and \methodb.
\item We propose \method, a novel MMV framework that is carefully designed to harness an inductive bias we identified in SOTA LVM-based forecasters, complemented by the strength of a numerical forecaster, with two technical variants \methoda\ and \methodb.
% \item We conduct comprehensive experiments on benchmark datasets to evaluate \method. The superior performance of \method\ over 14 SOTA baselines suggests a new paradigm for MMV-based time series learning.
\item We conduct comprehensive experiments on benchmark datasets to evaluate \method, demonstrating its superior performance over 14 SOTA baselines and highlighting its potential as a new paradigm for MMV-based time series learning.
\end{itemize}

\section{Related Work}\label{sec.relate}

To the best of our knowledge, this is the first work to explore LVMs in a decomposition-based MMV framework for LTSF. Our work relates to \textbf{LVM-based time series forecasting (TSF)}, \textbf{Multi-modal TSF}, and \textbf{Decomposition-based TSF}, which are discussed below.

\textbf{LVM-based TSF}. %Various vision models, such as \texttt{ResNet} \cite{he2016deep}, \texttt{VGG-Net} \cite{simonyan2014very}, ViT \cite{dosovitskiy2021image}, {\em etc.}, have been used for TSF \cite{zeng2023pixels,yang2024vitime}. Image-pre-trained CNNs have also been investigated in the past, such as pre-trained \texttt{ResNet} \cite{he2016deep}, \texttt{Inception-v1} \cite{szegedy2015going}, \texttt{VGG-19} \cite{simonyan2014very} for LTSF \cite{li2020forecasting}. LVM-based solution is an emergent subject. Most of them are developed for time series classification, such as \texttt{AST} \cite{gong2021ast} and \texttt{ViTST} \cite{li2023time}, which use \texttt{DeiT} \cite{touvron2021training} and \texttt{Swin} \cite{liu2021swin}, respectively. In contrast, TSF task has seen less efforts in using LVMs, possibly because LVMs are less adept at low-level tasks than high-level tasks. The most salient method is \texttt{VisionTS} \cite{chen2025visionts}, which adapts \texttt{MAE} \cite{he2022masked}, to zero-shot and few-shot TSF. Another work \texttt{ViTime} has pre-trained \texttt{ViT} from scratch using synthetic imaged time series for TSF, but hasn't explored knowledge transfer of image-pre-trained LVMs to TSF task. The above methods focus solely on vision models without integrating imaged time series with other views or modalities.%For detailed discussion about the existing literature on LVMs for time series analysis, we refer readers to \cite{ni2025harnessing}.
Various vision models, such as \texttt{ResNet} \cite{he2016deep}, \texttt{VGG-Net} \cite{simonyan2014very}, and \texttt{ViT} \cite{dosovitskiy2021image}, have been applied to TSF \cite{zeng2023pixels,yang2024vitime}, with some studies exploring image-pretrained CNNs like \texttt{ResNet} \cite{he2016deep}, \texttt{Inception-v1} \cite{szegedy2015going}, and \texttt{VGG-19} \cite{simonyan2014very} for LTSF \cite{li2020forecasting}. The use of LVMs in this area is still emerging, with most efforts focused on time series classification (e.g., \texttt{AST} \cite{gong2021ast} uses \texttt{DeiT} \cite{touvron2021training}, \texttt{ViTST} \cite{li2023time} uses \texttt{Swin} \cite{liu2021swin}). In contrast, LVMs have seen limited use in TSF, likely due to their lower effectiveness on low-level ({\em i.e.}, numerical-level) tasks. The most notable method is \texttt{VisionTS} \cite{chen2025visionts}, which adapts \texttt{MAE} \cite{he2022masked} for zero-shot and few-shot TSF. Another method, \texttt{ViTime} \cite{yang2024vitime}, trains \texttt{ViT} \cite{dosovitskiy2021image} from scratch on synthetic imaged time series but does not explore transferring knowledge from image-pretrained LVMs. Importantly, these approaches rely solely on vision models without incorporating other views or modalities.

\textbf{Multi-modal TSF.} %Recently, large vision-language models (VLMs), such as LLaVA \cite{liu2023visual}, CLIP \cite{radford2021learning}, ViLT \cite{kim2021vilt}, {\em etc.}, which involve pre-trained large vision encoders, have been explored for time series analysis \cite{wimmer2023leveraging,prithyani2024feasibility,zhuang2024see,zhong2025time}. The most pertinent is \texttt{Time-VLM} \cite{zhong2025time}, which builds a forecaster upon \texttt{ViLT} \cite{kim2021vilt} for encoding numerical and visual views of time series, along with some contextual texts. By integrating rich information with a large model, \texttt{Time-VLM} shows effectiveness in TSF. However, its view fusion strategy largely follows \texttt{ViLT} backbone. Without an in-depth design specific to time series, the fusion outcome is subject to be suboptimal.% For detailed discussion about the existing literature on VLMs for time series analysis, we refer readers to \cite{}.
Recently, large VLMs such as \texttt{LLaVA} \cite{liu2023visual}, \texttt{CLIP} \cite{radford2021learning}, and \texttt{ViLT} \cite{kim2021vilt} have been explored for time series analysis \cite{wimmer2023leveraging,prithyani2024feasibility,zhuang2024see,zhong2025time}. The most relevant is \texttt{Time-VLM} \cite{zhong2025time}, which builds a forecaster on \texttt{ViLT} \cite{kim2021vilt} to encode numerical and visual views, along with contextual texts. While integrating rich information with a large model, \texttt{Time-VLM} demonstrates promising results in TSF. However, its fusion strategy closely follows the \texttt{ViLT} backbone and lacks time-series-specific design, leading to potentially suboptimal performance.

\textbf{Decomposition-based TSF.} %Decomposition is a widely used technique in TSF. In particular, seasonal-trend decomposition (STD) has been used by popular models such as \texttt{Autoformer} \cite{wu2021autoformer}, \texttt{FEDformer} \cite{zhou2022fedformer}, and \texttt{DLinear} \cite{zeng2023transformers}. Recently, \texttt{Leddam} \cite{yu2024revitalizing} proposes a learnable decomposition kernel to substitute the moving-average kernel in STD. Residual decomposition is another technique. \texttt{N-BEATS} \cite{oreshkin2020n} uses it for alleviating forecasting errors. It inspires other works such as \texttt{DEPTS} \cite{fan2022depts} and \texttt{CycleNet} \cite{lin2024cyclenet} that adapt this technique for period-trend decomposition. \texttt{SparseTSF} \cite{lin2024sparsetsf} predicts periods but does not explicitly perform decomposition. \texttt{SSCNN} \cite{deng2024parsimony} introduces an attention-based decompostion to extract long-term, short-term, seasonal, and spatial components for TSF. However, these methods never explore LVMs. The proposed \method\ shares insights with residual decomposition, but is designed to harness an inductive bias of LVMs during TSF for adaptive decomposition, thus is remarkably different form the above methods.
% \texttt{DEPTS} \cite{} parameterizes periodicity and uses a residual decomposition that resembles \texttt{N-BEATS} \cite{}
Decomposition is a common technique in TSF, with seasonal-trend decomposition (STD) employed by models like \texttt{Autoformer} \cite{wu2021autoformer}, \texttt{FEDformer} \cite{zhou2022fedformer}, and \texttt{DLinear} \cite{zeng2023transformers}. Recent work such as \texttt{Leddam} \cite{yu2024revitalizing} replaces the traditional moving-average kernel in STD with a learnable one. Residual decomposition is another approach, used by \texttt{N-BEATS} \cite{oreshkin2020n} to reduce forecasting errors and later adopted by \texttt{DEPTS} \cite{fan2022depts} and \texttt{CycleNet} \cite{lin2024cyclenet} for period-trend decomposition. While \texttt{SparseTSF} \cite{lin2024sparsetsf} predicts periods without explicit decomposition, \texttt{SSCNN} \cite{deng2024parsimony} introduces an attention-based method to extract long-term, short-term, seasonal, and spatial components. However, none of these methods incorporate LVMs. In contrast, our proposed \method\ shares insights with residual decomposition but is uniquely designed to exploit the inductive bias of LVMs for adaptive decomposition, setting it apart from prior works.

% To sum up, the proposed \method\ is different from any of the above threads of methods, but integrates the benefits of pre-trained LVMs, MMV framework, and decomposition technique.
In summary, the proposed \method\ framework is distinct from existing approaches, yet integrates the strengths of pre-trained LVMs, the MMV framework, and decomposition techniques.

\section{Decomposition-Based Multi-Modal View (\method) Framework}\label{sec.method}

\begin{figure*}[t]
\centering
\includegraphics[width=0.97\linewidth]{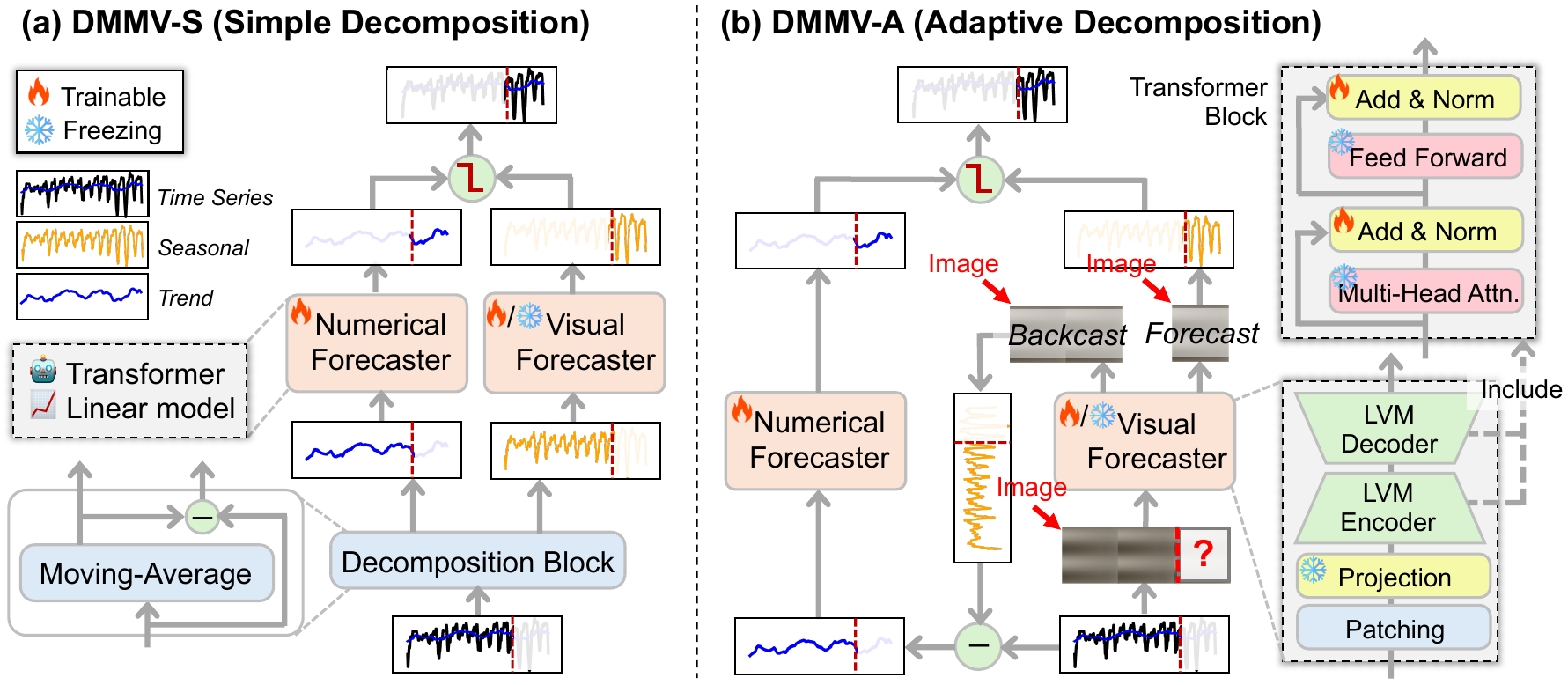}
% \vspace{-1em}
\caption{An overview of \method\ framework. (a) \methoda\ uses %a moving-average decomposition
moving-average to extract trend and seasonal components. (b) \methodb\ uses a backcast-residual decomposition to automatically learn trend and seasonal components. In (b), the gray blocks are gray-scale images. ``?'' marks masks.
%\yu{fig 1a: moving average part can be merged into the Decomposition Block; transformer and linear model part merged into forecaster block; fig 2b: visual forecaster part is confusing: not sure if it includes transformer block or not?; backcast and forecast blocks - meaning of `gray' block}
}\label{fig.method}
\vspace{-0.2cm}
\end{figure*}

% Firstly, a word about the LTSF problem\yu{?}.
\textbf{Problem Statement}. Given a multivariate time series (MTS) $\mat{X}=[\mat{x}^{1}, ..., \mat{x}^{D}]^{\top}\in\mathbb{R}^{D\times T}$ within a {\em look-back window} of length $T$, where $\mat{x}^{i}\in\mathbb{R}^{T}$ ($1\le i \le D)$ is a UTS of the $i$-th variate, the goal of LTSF is to estimate the most likely values of the MTS at future $H$ time steps, {\em i.e.}, $\mat{\hat{Y}}\in\mathbb{R}^{D\times H}$, such that the difference between the estimation and the ground truth $\mat{Y}=\mat{X}_{T+1:T+H}\in\mathbb{R}^{D\times H}$ is minimized in terms of mean squared error (MSE), {\em i.e.}, $\frac{1}{D\cdot H}\sum_{i=1}^{D}\sum_{t=1}^{H}\|\mat{\hat{Y}}_{it} - \mat{Y}_{it}\|_{2}^{2}$.%\yu{ $\frac{1}{D\times H}$ better?}

\begin{wrapfigure}{r}{0.5\textwidth}
% \begin{figure}[!t]
\centering
\includegraphics[width=0.5\textwidth]{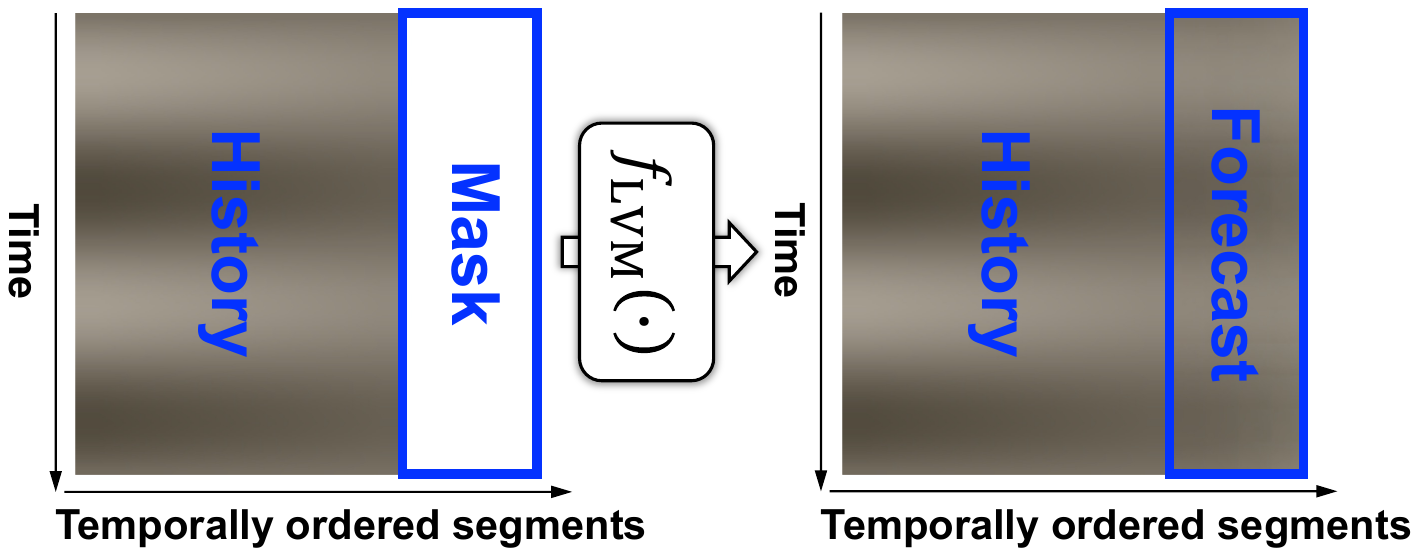}
% \vspace{-1em}
\caption{An illustration of an LVM forecaster}\label{fig.preliminary}
\vspace{-0.3cm}
% \end{figure}
\end{wrapfigure}

\textbf{Preliminaries}. Masked autoencoder (\texttt{MAE}) \cite{he2022masked} is pre-trained self-supervisedly by reconstructing masked image patches using ImageNet dataset \cite{deng2009imagenet}. To adapt it to LTSF, \texttt{VisionTS} \cite{chen2025visionts} adopts a period-based imaging technique introduced by \texttt{TimesNet} \cite{wu2023timesnet}. Specifically, each length-$T$ UTS $\mat{x}^{i}$ is segmented into $\lfloor T/P\rfloor$ subsequences of length $P$, where $P$ is set to be the period of $\mat{x}^{i}$, which can be obtained using Fast Fourier Transform (FFT) on $\mat{x}^{i}$ \cite{wu2023timesnet} or from prior knowledge on sampling frequency. The subsequences are stacked to form a 2D image $\mat{I}^{i}\in\mathbb{R}^{P\times\lfloor T/P\rfloor}$. After standard-deviation normalization, $\mat{I}^{i}$ is duplicated 3 times to form a gray image of size $P\times\lfloor T/P\rfloor\times 3$, followed by a bilinear interpolation to resize it to an image $\mat{\tilde{I}}^{i}$ %$\mat{\tilde{I}}^{i}\in\mathbb{R}^{224\times 224\times 3}$ that fits the 
of size $224\times 224\times 3$ to fit the input requirement of \texttt{MAE}. %LVMs.
As Fig. \ref{fig.preliminary} shows, the forecast %of $\mat{x}^{i}$
is achieved by reconstructing a right-appended masked area of $\mat{\tilde{I}}^{i}$, corresponding to the future horizon of $\mat{x}^{i}$. The forecast $\mat{\hat{y}}^{i}\in\mathbb{R}^{H}$ can be recovered from the reconstructed area by de-normalization and reverse transformation. The forecast of MTS $\mat{X}$ is achieved by forecasting over $\mat{x}^{1}$, ..., $\mat{x}^{D}$ in parallel, following the channel-independence assumption \cite{nie2023time}.

\begin{figure*}[!h]
\centering
\includegraphics[width=0.99\linewidth]{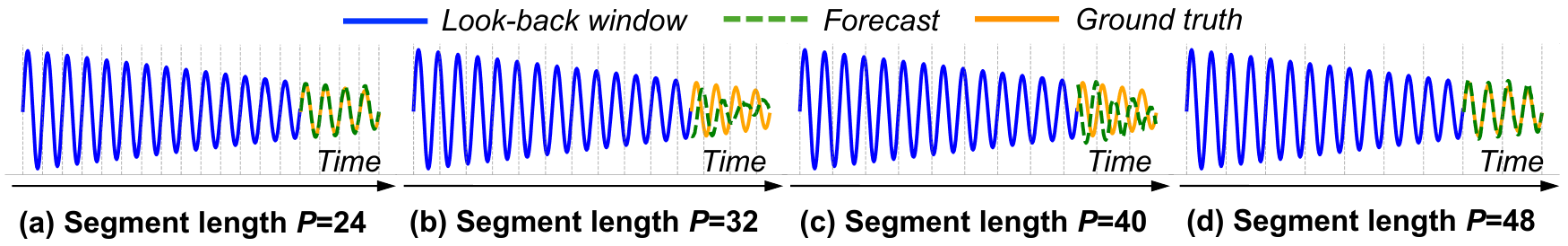}
% \vspace{-1em}
\caption{An illustration of LVM forecaster's inductive bias. The time series has a period of 24. The vertical dashed lines mark the segment points. The example indicates a bias towards segment lengths that are multiples of the period in (a)(d) over other segment lengths in (b)(c).}\label{fig.bias}
\vspace{-0.3cm}
\end{figure*}

\textbf{An Inductive Bias.} %\yu{Is it ok to keep this term as is, or should we replace it with a more descriptive alternative?@jingchao}.
% Because of the period-based imaging, along with the spatial consistency implicitly enforced during \texttt{MAE}'s pixel inferences, \texttt{VitionTS}'s forecasts are largely biased by {\em inter-period consistency}, overwhelming the global trend of the time series. Fig. \ref{fig.bias} shows \texttt{VisionTS}'s forecasts on a synthetic sinusoid time series of period 24. In Fig. \ref{fig.bias}(a)(b)(c)(d), the segmenting length $P$ is set to 24, 32, 40, 48, respectively. As can be seen, by shifting $P$ from $1\times$period to $2\times$period, the forecasts change from correct to wrong, then to correct, suggesting a strong inductive bias towards periodic patterns. However, the forecasts are not simply ``copying the past'' as can be seen from the decreasing intra-period amplitude in the forecasts, suggesting LVMs' ability in inferring intra-period, local trend.
Due to period-based imaging and the spatial consistency enforced during \texttt{MAE}'s pixel inference, \texttt{VitionTS} exhibits a strong bias toward {\em inter-period consistency}, often overshadowing the global trend. Fig. \ref{fig.bias} illustrates \texttt{VisionTS}'s forecasts on a synthetic sinusoidal time series with a period of 24. As shown in Fig. \ref{fig.bias}(a)-(d), where the segment length $P$ varies from 24 to 48, forecasts alternate between accurate and inaccurate as $P$ shifts from $1\times$period to $2\times$period, highlighting a strong inductive bias toward periodicity. Notably, the forecasts aren't mere repetitions -- the decreasing intra-period amplitude indicates that LVMs can still capture local trends within each period. More quantitative results are deferred to Appendix \ref{app. analysis of inductive bias}.

% Based on this observation, we design our MMV framework to harness the LVM's inductive bias and complement its limitation. The idea is to use a {\em visual forecaster} $f_{\text{vis}}(\cdot)$ ({\em i.e.}, LVM) to capture solely the periodic patterns from the {\em visual view}, easing its forecasting task, while use a {\em numerical forecaster} $f_{\text{num}}(\cdot)$ to capture the global trend patterns from the {\em numerical view} to strengthen the overall forecasting. % which is detailed next.% Unless otherwise noted, the superscript $i$ in the notations is omitted for brevity in the following.% due to the same operation for different variates.
% Fig. \ref{fig.method} is an overview of two variants of the proposed \method\ framework -- \methoda\ and \methodb. %which has a {\em numerical forecaster} $f_{\text{num}}(\cdot)$ and a {\em visual forecaster} $f_{\text{vis}}(\cdot)$ for the numerical view and visual view of the input time series, respectively.
% \method\ adopts a decomposition architecture. Unlike existing decomposition-based methods \cite{wu2021autoformer,lin2024sparsetsf,deng2024parsimony,lin2024cyclenet}, \method\ is such designed to accommodate the inductive bias of LVM forecasters as identified above.%we discovered from the SOTA \texttt{MAE}-based forecaster \cite{chen2025visionts}.% \yu{move this paragraph to section \ref{sec.methoda}}

Motivated by this observation, we design the \method\ framework to leverage the inductive bias of LVMs while addressing their limitations. Specifically, the {\em visual forecaster} $f_{\text{vis}}(\cdot)$ ({\em i.e.}, LVM) focuses on capturing periodic patterns from the {\em visual view}, while the {\em numerical forecaster} $f_{\text{num}}(\cdot)$ models global trends from the {\em numerical view}, resulting in more balanced forecasting. Fig. \ref{fig.method} presents the two \method\ variants -- \methoda\ and \methodb\ -- within a decomposition-based architecture. Unlike prior approaches \cite{wu2021autoformer,lin2024sparsetsf,deng2024parsimony,lin2024cyclenet}, \method\ is explicitly designed to align with the inductive bias of LVMs.

\subsection{\method\ with Simple Decomposition (\methoda)}\label{sec.methoda}

%The first instance of \method\, {\em i.e.},
\methoda\ adopts a simple moving-average (MOV) decomposition \cite{wu2021autoformer}, which explicitly decomposes an input time series $\mat{x}^{i}$ into a trend part and a seasonal (or periodic) part, reflecting the long-term progression and the seasonality of $\mat{x}^{i}$, respectively. Basically, MOV uses a kernel ({\em i.e.}, a sliding window) of length $2\lfloor P/2\rfloor+1$ to %smooth out periodic fluctuations and highlight the global trend,
extract the component with frequency lower than $\mat{x}^{i}$'s sampling frequency ({\em i.e.}, $1/P$), highlighting the global trend. The residual component is the seasonal part. This operation constitutes the decomposition block in Fig. \ref{fig.method}(a).
\begin{equation}\label{eq.moveavg}
\begin{aligned}
\mat{x}_{\text{trend}}^{i}=\texttt{Moving-Average}(\texttt{Padding}(\mat{x}^{i})),~~~\mat{x}_{\text{season}}^{i}=\mat{x}^{i} - \mat{x}_{\text{trend}}^{i},~~~1\le i\le D
\end{aligned}
\end{equation}
where $\texttt{Padding}(\cdot)$ keeps the length of $\mat{x}^{i}$ fixed.

% The visual forecaster $f_\text{vis}(\cdot)$ takes $\mat{x}_{\text{season}}^{i}$ as input, performs the transformation as aforementioned to produce a $224\times 224\times 3$ image $\mat{\tilde{I}}_{\text{season}}^{i}$, and output the forecast $\mat{\hat{y}}_{\text{season}}^{i}\in\mathbb{R}^{H}$ for the seasonal part. For the numerical forecaster $f_{\text{num}}(\cdot)$, instead of enforcing an inductive bias with a specialized design, we adopt a generalized forecaster that can flexibly encode long-term dependency. Specifically, we consider two options: (1) A simple linear forecaster, {\em i.e.}, $\mat{\hat{y}}_{\text{trend}}^{i}=f_{\text{num}}(\mat{x}_{\text{trend}}^{i})=\mat{W}\mat{x}_{\text{trend}}^{i}+\mat{b}$, where $\mat{W}\in\mathbb{R}^{T\times H}$ and $\mat{b}\in\mathbb{R}^{H}$ are weight and bias, respectively. This is inspired by the power of linear models in LTSF as validated by \cite{zeng2023transformers,lin2024sparsetsf,lin2024cyclenet}; (2) A Transformer-based forecaster following \texttt{PatchTST} \cite{nie2023time}, which divides $\mat{x}_{\text{trend}}^{i}$ into $N$ length-$L$ patches, where $N=\lfloor T/L\rfloor + 1$, stacks them to form $\mat{X}_{\text{trend}}^{i}\in\mathbb{R}^{L\times N}$, and performs
The visual forecaster $f_\text{vis}(\cdot)$ transforms the input $\mat{x}_{\text{season}}^{i}$ into a $224\times 224\times 3$ image $\mat{\tilde{I}}_{\text{season}}^{i}$, and outputs the forecast $\mat{\hat{y}}_{\text{season}}^{i}\in\mathbb{R}^{H}$ for the seasonal component. For the numerical forecaster $f_{\text{num}}(\cdot)$, rather than imposing a specialized inductive bias, we adopt a general-purpose architecture capable of capturing long-term dependencies. We investigate the feasibility of two options and leave other explorations as a future work: (1) A simple linear model motivated by the proven effectiveness of linear methods in LTSF \cite{zeng2023transformers,lin2024sparsetsf,lin2024cyclenet}, {\em i.e.}, $\mat{\hat{y}}_{\text{trend}}^{i}=f_{\text{num}}(\mat{x}_{\text{trend}}^{i})=\mat{W}\mat{x}_{\text{trend}}^{i}+\mat{b}$, where $\mat{W}\in\mathbb{R}^{H\times T}$ and $\mat{b}\in\mathbb{R}^{H}$ are weight and bias, respectively; and (2) A Transformer-based model inspired by \texttt{PatchTST} \cite{nie2023time}, which segments $\mat{x}_{\text{trend}}^{i}$ into $N$ length-$L$ patches, where $N=\lfloor T/L\rfloor + 1$, to form the input $\mat{X}_{\text{trend}}^{i}\in\mathbb{R}^{L\times N}$, and performs
\begin{equation}\label{eq.transformer}
\small
\begin{aligned}
\mat{\tilde{X}}_{\text{trend}}^{i}=\mat{W}_{\text{pro}}\mat{X}_{\text{trend}}^{i}+\mat{W}_{\text{pos}}\rightarrow\mat{\hat{X}}_{\text{trend}}^{i}=\texttt{Transformer}(\mat{\tilde{X}}_{\text{trend}}^{i})\rightarrow\mat{\hat{y}}_{\text{trend}}^{i}=\texttt{Linear}(\texttt{Flatten}(\mat{\hat{X}}_{\text{trend}}^{i}))
\end{aligned}
\end{equation}
to achieve the forecast $\mat{\hat{y}}_{\text{trend}}^{i}\in\mathbb{R}^{H}$ for the trend part, where $\mat{W}_{\text{pro}}\in\mathbb{R}^{D'\times L}$ is the weight to project the patches to $D'$-dimensional embeddings, $\mat{W}_{\text{pos}}\in\mathbb{R}^{D'\times N}$ is a learnable positional encoding, $\texttt{Flatten}(\cdot)$ and $\texttt{Linear}(\cdot)$ are flatten and linear operators.

Finally, $\mat{\hat{y}}_{\text{season}}^{i}$ and $\mat{\hat{y}}_{\text{trend}}^{i}$ are merged to produce the overall forecast $\mat{\hat{y}}^{i}$ for the $i$-th variate. In particular, instead of using the regular summation-based merge, we design an adaptive merge function with a {\em light-weight gate} $g=\texttt{sigmoid}(w_{g})\in [0, 1]$, where $w_{g}$ is a learnable scalar parameter. To sum up, the overall process of \methoda\ is as follows.
\begin{equation}\label{eq.methoda}
\begin{aligned}
\mat{\hat{y}}^{i}=g\circ\mat{\hat{y}}_{\text{season}}^{i}+(1-g)\circ\mat{\hat{y}}_{\text{trend}}^{i},~~~\text{where}~~\mat{\hat{y}}_{\text{season}}^{i}=f_{\text{vis}}(\mat{\tilde{I}}_{\text{season}}^{i}),~~~\mat{\hat{y}}_{\text{trend}}^{i}=f_{\text{num}}(\mat{x}_{\text{trend}}^{i})
\end{aligned}
\end{equation}

\textbf{Remark}. One limitation of \methoda\ is the explicit trend-seasonal decomposition placed on the input $\mat{x}^{i}$, which will enforce $f_{\text{num}}(\cdot)$  and $f_{\text{vis}}(\cdot)$ to fit \textbf{pre-defined components} extracted by a certain kernel size. This is not flexible and may not fully leverage LVMs' potential. To address it, we develop \methodb\ to have an adaptive decomposition in the next.

\subsection{\method\ with Adaptive Decomposition (\methodb)}\label{sec.method.adaptive}

\begin{wrapfigure}{r}{0.47\textwidth}
% \begin{figure}[!t]
\centering
\includegraphics[width=0.47\textwidth]{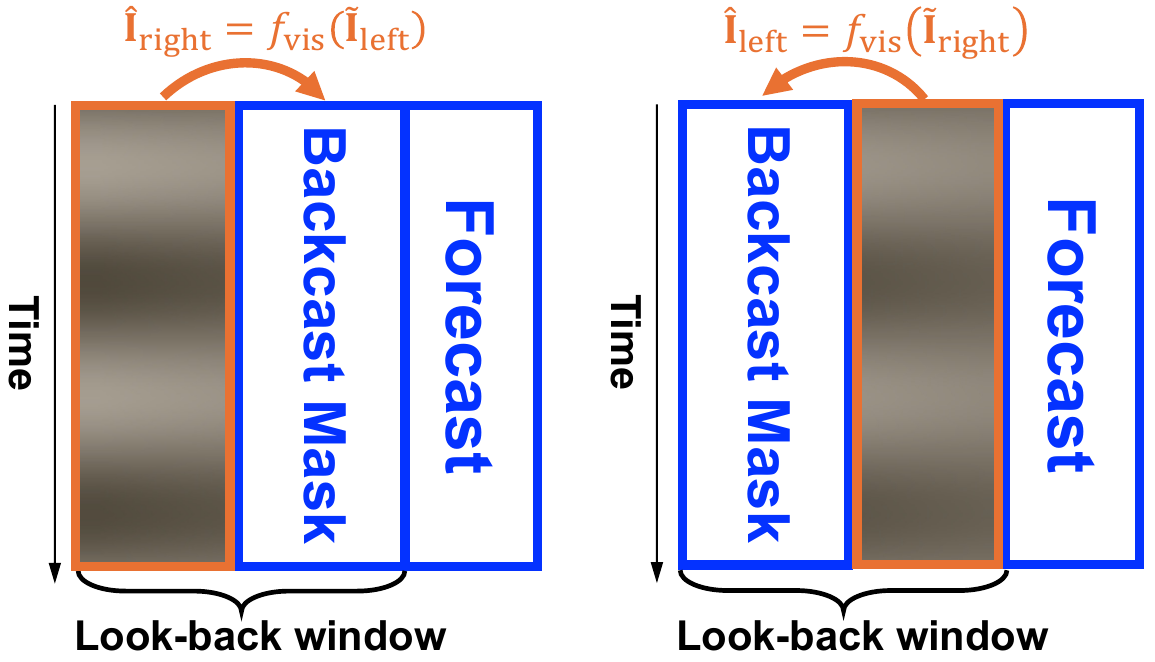}
\caption{An illustration of \maskmethod.}\label{fig.mask}
\vspace{-0.8cm}
% \end{figure}
\end{wrapfigure}

Unlike \methoda, \methodb\ {\em implicitly} decomposes the input $\mat{x}^{i}$ into trend and seasonal components tailored to the strengths of the numerical and visual forecasters, respectively. This is achieved via a {\em backcast-residual} mechanism (Fig. \ref{fig.method}(b)) that leverages LVMs' bias toward periodic patterns. The input $\mat{x}^{i}$ is first transformed into an image $\mat{\tilde{I}}^{i}$ using period-based imaging. Before forecasting, $f_{\text{vis}}(\cdot)$ is used to {\em backcast} the look-back window by reconstructing masked segments of $\mat{\tilde{I}}^{i}$. An effective masking strategy must: (1) enable full-window reconstruction; (2) align with the forecasting setup (Fig. \ref{fig.preliminary}); and (3) minimize the usage of $f_{\text{vis}}(\cdot)$ to avoid computational overhead. To meet these criteria, we propose an efficient {\em BackCast-Masking} (\maskmethod) strategy (Fig. \ref{fig.mask}), which applies two passes: masking and reconstructing the left and right halves of $\mat{\tilde{I}}^{i}$, respectively.
\begin{equation}\label{eq.mask}
\begin{aligned}
\mat{\hat{I}}^{i}=[\mat{\hat{I}}_{\text{left}}^{i}; \mat{\hat{I}}_{\text{right}}^{i}],~~~\text{with}~~\mat{\hat{I}}_{\text{left}}^{i}=f_{\text{vis}}(\mat{\tilde{I}}_{\text{right}}^{i}),~~~\mat{\hat{I}}_{\text{right}}^{i}=f_{\text{vis}}(\mat{\tilde{I}}_{\text{left}}^{i})
\end{aligned}
\end{equation}
where $\mat{\tilde{I}}_{\text{right}}^{i}$ ($\mat{\tilde{I}}_{\text{left}}^{i}$) is the masked image with right (left) area unmasked, $\mat{\hat{I}}_{\text{left}}^{i}$ ($\mat{\hat{I}}_{\text{right}}^{i}$) is the reconstructed left (right) area, $\mat{\hat{I}}^{i}$ is the reconstruction, or backcast, of $\mat{\tilde{I}}^{i}$ by concatenating $\mat{\hat{I}}_{\text{left}}^{i}$ and $\mat{\hat{I}}_{\text{right}}^{i}$.

% \maskmethod\ meets the three desiderata since (1) the full area of $\mat{\tilde{I}}^{i}$ is predicted; (2) a continuous set of segments in $\mat{\tilde{I}}_{\text{right}}^{i}$ ($\mat{\tilde{I}}_{\text{left}}^{i}$) is used to predict its contiguous set of segments $\mat{\hat{I}}_{\text{left}}^{i}$ ($\mat{\hat{I}}_{\text{right}}^{i}$), resembling the forecasting process in Fig. \ref{fig.preliminary}; (3) the twice use of $f_{\text{vis}}(\cdot)$ is minimum considering the fact that at least a part of $\mat{\tilde{I}}^{i}$ needs to be unmasked to predict an masked area, and the unmasked area should be masked and predicted later for reconstructing the full look-back window in $\mat{\tilde{I}}^{i}$.

\maskmethod\ satisfies all three criteria: (1) it enables full reconstruction of $\mat{\tilde{I}}^{i}$; (2) it uses contiguous segments in $\mat{\tilde{I}}_{\text{right}}^{i}$ ($\mat{\tilde{I}}_{\text{left}}^{i}$) to predict adjacent segments $\mat{\hat{I}}_{\text{left}}^{i}$ ($\mat{\hat{I}}_{\text{right}}^{i}$), mirroring the forecasting process in Fig. \ref{fig.preliminary}; and (3) it minimizes the use of $f_{\text{vis}}(\cdot)$ -- only two passes are needed, as some unmasked regions of $\mat{\tilde{I}}^{i}$ are required for prediction and must later be masked to complete the full reconstruction.

% It is noteworthy that the backcast in image $\mat{\hat{I}}^{i}$ is biased towards the periodic patterns in $\mat{\tilde{I}}^{i}$. By de-normalization and reverse transformation, a backcast time series $\mat{\hat{x}}^{i}\in\mathbb{R}^{T}$ can be recovered from $\mat{\hat{I}}^{i}$, which is biased towards the periodic patterns in $\mat{x}^{i}$. Thus the residual $\Delta\mat{x}^{i}=\mat{x}^{i}-\mat{\hat{x}}^{i}$ is biased towards the trend component. Inspired by this, as shown in Fig. \ref{fig.method}(b), we use $f_{\text{num}}(\cdot)$ to process $\Delta\mat{x}^{i}$, resembling $f_{\text{num}}(\cdot)$'s role in \methoda, and denote the output as $\mat{\hat{y}}_{\text{trend}}^{i}\in\mathbb{R}^{H}$. Also, we use $f_{\text{vis}}(\cdot)$ to perform forecasting based on $\mat{\tilde{I}}^{i}$, which tends to produce a seasonal forecast. Hence we denote its output as $\mat{\hat{y}}_{\text{season}}^{i}\in\mathbb{R}^{H}$. Finally, $\mat{\hat{y}}_{\text{trend}}^{i}$ and $\mat{\hat{y}}_{\text{season}}^{i}$ are merged using the same gating mechanism as in Eq.~\eqref{eq.methoda}. To sum up, the overall process of \methodb\ is as follows.

Notably, the backcast in image $\mat{\hat{I}}^{i}$ is biased toward the periodic patterns in $\mat{\tilde{I}}^{i}$. After de-normalization and reverse transformation, a backcast time series $\mat{\hat{x}}^{i}\in\mathbb{R}^{T}$ is recovered, reflecting periodic component in $\mat{x}^{i}$. The residual $\Delta\mat{x}^{i}=\mat{x}^{i}-\mat{\hat{x}}^{i}$ therefore emphasizes the trend. As shown in Fig. \ref{fig.method}(b), we feed $\Delta\mat{x}^{i}$ into $f_{\text{num}}(\cdot)$ to produce $\mat{\hat{y}}_{\text{trend}}^{i}\in\mathbb{R}^{H}$, analogous to its role in \methoda. Meanwhile, $f_{\text{vis}}(\cdot)$ predicts from $\mat{\tilde{I}}^{i}$, likely yielding the forecast of seasonal component $\mat{\hat{y}}_{\text{season}}^{i}\in\mathbb{R}^{H}$. Finally, $\mat{\hat{y}}_{\text{trend}}^{i}$ and $\mat{\hat{y}}_{\text{season}}^{i}$ are fused via the same gating mechanism as Eq.~\eqref{eq.methoda}. In summary, this defines the overall process of \methodb\ as follows.
\begin{equation}\label{eq.methodb}
\begin{aligned}
\mat{\hat{y}}^{i}=g\circ\mat{\hat{y}}_{\text{season}}^{i}+(1-g)\circ\mat{\hat{y}}_{\text{trend}}^{i},~~~\text{where}~~\mat{\hat{y}}_{\text{season}}^{i}=f_{\text{vis}}(\mat{\tilde{I}}^{i}),~~~\mat{\hat{y}}_{\text{trend}}^{i}=f_{\text{num}}(\Delta\mat{x}^{i})
\end{aligned}
\end{equation}

\textbf{Remark}. %Compared to \methoda, \methodb\ automatically learns a decomposition of $\mat{x}^{i}$ that best fits $f_{\text{num}}(\cdot)$ and $f_{\text{vis}}(\cdot)$ to the forecasting task. As we evaluated in $\S$\ref{sec.exp.analysis}, such an adaptive decomposition indeed learns the (adaptive) seasonal and trend parts of $\mat{x}^{i}$ due to the inductive bias of $f_{\text{vis}}(\cdot)$. In contrast to the backcast mechanism in \texttt{N-BEATS} \cite{oreshkin2020n}, where the goal is to merely extract the backcast errors of an input time series, our method is tailored to the purpose of harnessing the inductive bias of LVM forecasters, thus is remarkably different.
Unlike \methoda, \methodb\ automatically learns a decomposition of $\mat{x}^{i}$ that optimally aligns $f_{\text{num}}(\cdot)$ and $f_{\text{vis}}(\cdot)$ with the forecasting task. As shown in $\S$\ref{sec.exp.analysis}, this adaptive decomposition effectively separates seasonal and trend components, leveraging the inductive bias of $f_{\text{vis}}(\cdot)$. Unlike the backcast %mechanism
in \texttt{N-BEATS} \cite{oreshkin2020n} -- designed merely to extract predictive errors -- our approach is specifically tailored to exploit LVMs' bias toward periodic patterns, making it fundamentally different.

\vspace{-0.2cm}

\subsection{Model Optimization}\label{sec.method.training}

% After obtaining $\mat{\hat{Y}}=[\mat{\hat{y}}^{1}, ..., \mat{\hat{y}}^{D}]^{\top}\in\mathbb{R}^{D\times H}$, \method\ is trained by minimizing the MSE between $\mat{\hat{Y}}$ and $\mat{Y}$ on training datasets, {\em i.e.}, $\frac{1}{D\cdot H}\sum_{i=1}^{D}\sum_{t=1}^{H}\|\mat{\hat{Y}}_{it} - \mat{Y}_{it}\|_{2}^{2}$. As illustrated in Fig. \ref{fig.method}, $f_{\text{num}}(\cdot)$ is fully trained from scratch, while $f_{\text{vis}}(\cdot)$ adopts pre-trained parameters in an LVM and is partially fine-tuned. We find fine-tuning only the norm layers give the best performance, which is consistent with the findings in \cite{zhou2023one}. For the choice of LVM, we tested two self-supervisedly pre-trained LVMs, {\em i.e.}, \texttt{MAE} \cite{he2022masked} and \texttt{SimMIM} \cite{xie2022simmim}, in $\S$\ref{sec.exp.ablation} and find \texttt{MAE} is better thus is set as default. The whole training process starts with freezing the entire $f_{\text{vis}}(\cdot)$ while training $f_{\text{num}}(\cdot)$ until a pre-defined number of epochs ({\em i.e.}, \red{10?}), and then unfreezing the norm layers of $f_{\text{vis}}(\cdot)$ and fine-tuning them together with $f_{\text{num}}(\cdot)$ until the maximum number of epochs or early stopping.% The training process is summarized in Algorithm \ref{} in Appendix \ref{}.

After obtaining $\mat{\hat{Y}}=[\mat{\hat{y}}^{1}, ..., \mat{\hat{y}}^{D}]^{\top}\in\mathbb{R}^{D\times H}$, \method\ is trained by minimizing the MSE between $\mat{\hat{Y}}$ and $\mat{Y}$, {\em i.e.}, $\frac{1}{D\cdot H}\sum_{i=1}^{D}\sum_{t=1}^{H}\|\mat{\hat{Y}}_{it} - \mat{Y}_{it}\|_{2}^{2}$. As shown in Fig. \ref{fig.method}, $f_{\text{num}}(\cdot)$ is trained from scratch, while $f_{\text{vis}}(\cdot)$ uses pre-trained LVM weights with partial fine-tuning. We find that fine-tuning only the normalization layers yields the best performance, consistent with the findings in \cite{zhou2023one}. For the choice of LVM, we tested \texttt{MAE} \cite{he2022masked} and \texttt{SimMIM} \cite{xie2022simmim}, both are self-supervisedly pre-trained LVMs, in $\S$\ref{sec.exp.ablation}. \texttt{MAE} performs better and is set as the default. Training begins with $f_{\text{vis}}(\cdot)$ frozen while $f_{\text{num}}(\cdot)$ is trained for a number of epochs ({\em e.g.}, $\sim$30). Then, the norm layers of $f_{\text{vis}}(\cdot)$ are unfrozen and fine-tuned jointly with $f_{\text{num}}(\cdot)$ until convergence or early stopping.

\section{Experiments}\label{sec.exp}

\vspace{-0.25cm}

In this section, we compare \method\ with the SOTA methods on LTSF benchmark datasets, and analyze its effectiveness with both quantitative and qualitative studies.

\textbf{Datasets}. %We use the widely adopted 8 benchmark datasets including ETT (Electricity Transformer Temperature) \cite{zhou2021informer}, encompassing ETTh1, ETTh2, ETTm1, ETTm2, Weather \cite{wu2021autoformer}, Illiness \cite{wu2021autoformer}, Traffic \cite{wu2021autoformer}, and Electricity \cite{trindade2015electricityloaddiagrams20112014}. All of the time series are MTS. We follow the standard evaluation protocol \cite{wu2021autoformer} and split all datasets into training/validation/test set in chronological order by the ratio 60\%/20\%/20\% for the ETT dataset and 70\%/10\%/20\% for the other datasets. The prediction length $H$ is set to \{24, 36, 48, 60\} for Illiness dataset and \{96, 192, 336, 720\} for other datasets. The look-back window length $T$ is set to 336. We defer detailed data descriptions to Appendix \ref{app. benchmark datasets}.
We adopt 8 widely used MTS benchmarks: ETT (Electricity Transformer Temperature) \cite{zhou2021informer}, including ETTh1, ETTh2, ETTm1, ETTm2; Weather \cite{wu2021autoformer}, Illness \cite{wu2021autoformer}, Traffic \cite{wu2021autoformer}, and Electricity \cite{trindade2015electricityloaddiagrams20112014}. Following standard protocols \cite{wu2021autoformer}, we split the datasets chronologically into training/validation/test sets using a 60\%/20\%/20\% ratio for ETT and 70\%/10\%/20\% for the others. The prediction horizon $H$ is set to \{24, 36, 48, 60\} for Illness, and \{96, 192, 336, 720\} for the remaining datasets. %The look-back window $T$ is fixed at 336.
By default, look-back window $T$ is 336. Full dataset details are provided in Appendix \ref{app. benchmark datasets}.

\textbf{The Compared Methods}. We compare \method\ with the SOTA methods, including VLM-based multi-modal model: (1) \texttt{Time-VLM} \cite{zhong2025time}; LVM-based model: (2) \texttt{VisionTS} \cite{chen2025visionts}; LLM-based models: (3) \texttt{Time-LLM} \cite{jin2023time}, (4) \texttt{GPT4TS} \cite{zhou2023one}, (5) \texttt{CALF} \cite{liu2025calf}; Transformer-based models: (6) \texttt{PatchTST} \cite{nie2023time}, (7) \texttt{FEDformer} \cite{zhou2022fedformer}, (8) \texttt{Autoformer} \cite{wu2021autoformer}, (9) \texttt{Stationary} \cite{liu2022non}, (10) \texttt{ETSformer} \cite{woo2022etsformer}, (11) \texttt{Informer} \cite{zhou2021informer}; and non-Transformer models: (12) \texttt{DLinear} \cite{zeng2023transformers}, (13) \texttt{TimesNet} \cite{wu2023timesnet}, (14) \texttt{CycleNet} \cite{lin2024cyclenet}. %\red{As we use the standard evaluation protocol, we collect the results of \texttt{Time-VLM} from \cite{zhong2025time}, the results of LLM-based models reproduced by \cite{tan2024language}, the results of Transformer-based models, \texttt{PatchTST} and \texttt{TimesNet} from \cite{chen2025visionts}, and the results of \texttt{CycleNet} from \cite{lin2024cyclenet}.} \red{Since the results of \texttt{VisionTS} in \cite{chen2025visionts} use look-back window length $T=1000$, which is different from the standard $T=336$. Thus we run \texttt{VisionTS} on all datasets with $T=336$ for fair comparison.}
As we use the standard evaluation protocol, we collect results from prior works: \texttt{Time-VLM} (\cite{zhong2025time}), LLM-based models (reproduced by \cite{tan2024language}), Transformer-based models, \texttt{PatchTST} and \texttt{TimesNet} (\cite{chen2025visionts}), and \texttt{CycleNet} (\cite{lin2024cyclenet}). Since \texttt{VisionTS} in \cite{chen2025visionts} originally uses dynamic look-back windows such as $T=1152$ and $T=2304$ for different datasets, we re-run it with $T=336$. %for fair comparison. % More details are in Appendix \ref{app. baseline}.
\texttt{CycleNet}'s results on the Illness dataset is unavailable in \cite{lin2024cyclenet}. Thus we run its official code on the Illness dataset.

% For the proposed \method, we include both \methoda\ and \methodb. By default, $f_{\text{num}}(\cdot)$ is set as a linear forecaster as stated in $\S$\ref{sec.methoda}, $f_{\text{vis}}(\cdot)$ is set as \texttt{MAE}. Moreover, we compare different variants of \method\ in the ablation analysis ($\S$\ref{sec.exp.ablation}), including setting $f_{\text{num}}(\cdot)$ as a Transformer forecaster and $f_{\text{num}}(\cdot)$ as \texttt{SimMIM}. For both \texttt{VisionTS} and \method, following \cite{chen2025visionts}, the period $P$ in the imaging technique is set according the each dataset's sampling frequency (see Table \ref{tab. dataset details} in Appendix \ref{app. baseline}). More details about the compared methods can be found in Appendix \ref{app. baseline}.

We evaluate both \method\ variants -- \methoda\ and \methodb. A linear forecaster ($\S$\ref{sec.methoda}) and \texttt{MAE} are set as the default in $f_{\text{num}}(\cdot)$ and $f_{\text{vis}}(\cdot)$, respectively. Ablation studies ($\S$\ref{sec.exp.ablation}) include variants with a Transformer-based $f_{\text{num}}(\cdot)$ and \texttt{SimMIM} as $f_{\text{vis}}(\cdot)$. Following \cite{chen2025visionts}, the imaging period $P$ ($\S$\ref{sec.method}) for both \texttt{VisionTS} and \method\ is set based on each dataset’s sampling frequency (see Appendix \ref{app. benchmark datasets}). Additional details on all compared methods are in Appendix \ref{app. baseline}.

\textbf{Evaluation}. Following \cite{nie2023time,zeng2023transformers,tan2024language}, we use Mean Squared Error (MSE) and Mean Absolute Error (MAE) to evaluate the LTSF performance of the compared methods.

\begin{table*}[!t]
\centering
\caption{%LTSF performance of the compared methods on the benchmark datasets.
LTSF performance comparison on the benchmark datasets. Lower MSE and MAE indicate better performance. \textbf{\textcolor{red}{Red}} values indicate the best MSE and MAE per row. %\texttt{Time-VLM} and \texttt{CycleNet} didn't report results on the Illness dataset thus their results are marked by ``--''.% The full results of all 16 compared methods can be found in Table \ref{tab. full baseline results} in Appendix \ref{app. full baseline results}.
\texttt{Time-VLM}'s results on the Illness dataset are unavailable in \cite{zhong2025time} and its code was unavailable at the time of this experiment.
}\label{tab.benchmark}
\vspace{0.2cm}
\tiny
% \resizebox{\textwidth}{!}{
\setlength{\tabcolsep}{1.9pt}{
\begin{tabular}{cc|cc|cc|cc|cc|cc|cc|cc|cc|cc|cc}
\toprule[1pt]
\multicolumn{2}{l}{\textbf{View}}          & \multicolumn{4}{c}{\textbf{Multi-Modal}}                                                                                                                                                        & \multicolumn{2}{c}{\textbf{Visual}}                                                           & \multicolumn{4}{c}{\textbf{Language}}                                                                                                         & \multicolumn{10}{c}{\textbf{Numerical}}\\ \cmidrule(lr){3-6} \cmidrule(lr){7-8} \cmidrule(lr){9-12} \cmidrule(lr){13-22}
\multicolumn{2}{l}{\textbf{Model}}                & \multicolumn{2}{c}{\methodb}                                            & \multicolumn{2}{c}{Time-VLM}                                                              & \multicolumn{2}{c}{VisionTS}                                                   & \multicolumn{2}{c}{GPT4TS}                                                                & \multicolumn{2}{c}{Time-LLM}                  & \multicolumn{2}{c}{PatchTST} & \multicolumn{2}{c}{CycleNet}                                                              & \multicolumn{2}{c}{TimesNet} & \multicolumn{2}{c}{DLinear}                & \multicolumn{2}{c}{FEDformer} \\
\end{tabular}
}
\vspace{-0.1cm}
% \yu{1. method level avg (optional) 2. two best baselines each have 10 missing results (illness dataset) which seems unfair...}
\end{table*}

\subsection{Experimental Results}\label{sec.exp.results}

Table \ref{tab.benchmark} summarizes the LTSF performance of 10 representative methods across four categories: MMV-based, visual-view-based, language-view-based, and numerical-view-based approaches, with full results for all 16 methods provided in Appendix \ref{app. full baseline results}. %The results of \texttt{Time-VLM} and \texttt{CycleNet} on the Illness dataset are marked by ``--'' since their results are not reported in \cite{zhong2025time} and \cite{lin2024cyclenet}.
\texttt{Time-VLM}'s results on the Illness dataset are not reported in \cite{zhong2025time} and its code is unavailable at the time of this experiment, thus are marked by ``--''. For \method, the stronger variant, \methodb, is reported. In Table \ref{tab.benchmark}, several key insights emerge: (1) MMV and visual-view methods generally outperform language-view methods, underscoring the effectiveness of LVMs, particularly when integrated within MMV frameworks; (2) Numerical-view models such as \texttt{PatchTST} and \texttt{CycleNet} remain competitive, especially on datasets where \texttt{VisionTS} underperforms ({\em e.g.}, ETTm2 and Electricity), highlighting their potential to complement visual models; (3) The strong results of \texttt{CycleNet}, a lightweight model with learnable decomposition, demonstrate the value of combining simplicity with structure in LTSF; (4) Notably, \methodb, which unifies visual and numerical views through a novel adaptive decomposition, outperforms the baselines in most cases, achieving 43 first-places and confirming its effectiveness; (5) Lastly, while \texttt{VisionTS} performs well on highly periodic datasets ({\em e.g.}, ETTh1, ETTm1, Traffic) due to \texttt{MAE}’s inductive bias toward periodicity, \methodb\ alleviates this bias, resulting in more generalizable forecasts.

% Moreover, Fig. \ref{fig.cdrank} provides critical difference diagrams \cite{han2022adbench} on the average rank of all 16 compared methods in terms of MSE and MAE overall prediction lengths on the 8 benchmark datasets. From Fig. \ref{fig.cdrank}, we observe \methoda\ is ranked 4.5/16 and 7.1/16 on MSE and MAE, respectively. Its inferior rank to \methodb\ validates the importance of the adaptive decomposition in Fig. \ref{fig.method}(b). However, \methoda's comparable ranks to \texttt{CycleNet} indicate even with a rigid moving-average decomposition, \methoda\ exhibits an ability that a SOTA powerful model with learnable decomposition has. Later, in $\S$\ref{}, we will analyze the difference between \methoda\ and \methodb\ in detail.

\begin{figure*}[!t]
\centering
\includegraphics[width=0.95\linewidth]{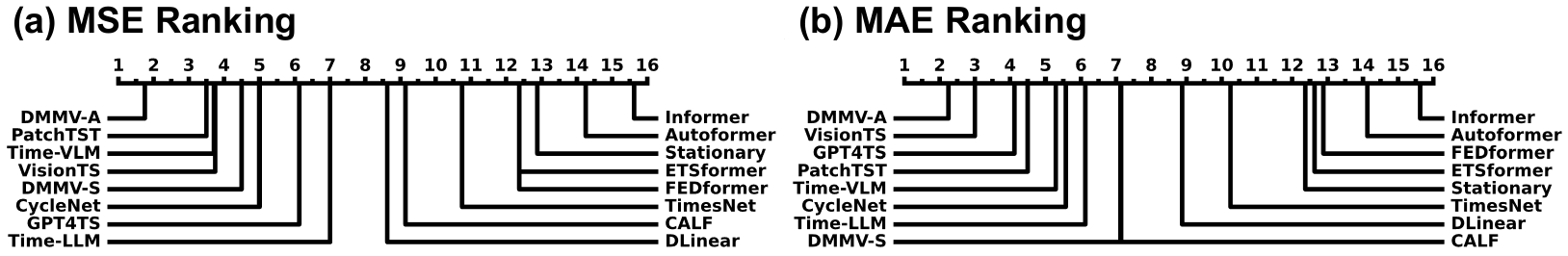}
% \vspace{-1em}
\caption{Critical difference (CD) diagram on the average rank of all 16 compared methods in terms of (a) MSE and (b) MAE over all benchmark datasets. The lower rank (left of the scale) is better.}\label{fig.cdrank}
\vspace{-0.3cm}
\end{figure*}

Fig. \ref{fig.cdrank} presents critical difference (CD) diagrams \cite{han2022adbench} showing the average rank of all 16 methods based on MSE and MAE across prediction lengths and all datasets. \methoda\ ranks 4.5/16 in MSE and 7.1/16 in MAE, underscoring the benefit of the adaptive decomposition used in \methodb\ (Fig. \ref{fig.method}(b)). From Fig. \ref{fig.cdrank}, \methoda's comparable ranks to \texttt{CycleNet} indicate even with a simpler, fixed decomposition, \methoda\ exhibits an ability that a strong SOTA model with learnable decomposition has. In $\S$\ref{sec.exp.analysis}, 
%Despite its simpler, fixed decomposition, from Fig. \ref{fig.cdrank}, \methoda\ exhibits an ability comparable to \texttt{CycleNet}, a strong SOTA model with learnable decomposition.
a detailed comparison between \methoda\ and \methodb\ is provided.

\begin{table*}[!t]
\centering
\caption{Ablation analysis of \methodb. MSE and MAE are averaged over different prediction lengths. Lower MSE and MAE are better. ``Improvement'' of each ablation is relative to \methodb.}\label{tab.ablation}
\vspace{0.2cm}
\scriptsize
\setlength{\tabcolsep}{5pt}{
\begin{tabular}{l|cc|cc|cc|cc}
\toprule[1pt]
\textbf{Dataset ($\rightarrow$)} & \multicolumn{2}{c|}{ETTh1}                                                     & \multicolumn{2}{c|}{ETTm1}                                                     & \multicolumn{2}{c|}{Illness}                                                   & \multicolumn{2}{c}{Weather} \\ \hline %\cmidrule{1-3} \cmidrule{4-5} \cmidrule{6-7} \cmidrule{8-9}
\textbf{Method ($\downarrow$),~~~Metric ($\rightarrow$)}  & MSE                                   & MAE                                   & MSE                                   & MAE                                   & MSE                                   & MAE                                   & MSE                                   & MAE \\ \midrule
\methodb\                 & 0.395                                 & \textbf{0.414}   & 0.340                                 & \textbf{0.371} & \textbf{1.407} & \textbf{0.771} & \textbf{0.217} & \textbf{0.256} \\ \hline
(a) $f_{\text{num}}(\cdot)\rightarrow$ \texttt{Transformer}         & 0.407                                 & 0.421                                 & 0.339                                 & 0.372                                 & 1.442                                 & 0.786                                 & 0.219                                 & 0.260  \\
~~~~~~Improvement & \red{-3.04\%} & \red{-1.69\%} & +0.29\% & \red{-0.27\%} & \red{-2.49\%} & \red{-1.95\%} & \red{-0.92\%} & \red{-1.56\%}\\ \hline
(b) $f_{\text{vis}}(\cdot)\rightarrow$ \texttt{SimMIM}        &    0.407        &0.415         &   0.345      &    0.377     &    1.649  &   0.814        &       0.227     & 0.261 \\
~~~~~~Improvement & \red{-3.04\%} & \red{-0.24\%} &\red{-1.47\%}  &\red{-1.62\%} &\red{-17.20\%}  &\red{-5.58\%}  &\red{-4.61\%}  &\red{-1.95\%} \\ \hline
(c) Gate $\rightarrow$ Sum        & 0.414                                 & 0.427                                 & 0.352                                 & 0.383                                 & 1.606                                 & 0.863                                 & 0.233                                 & 0.278 \\
~~~~~~Improvement & \red{-4.81\%} &\red{-3.14\%}  & \red{-3.53\%} & \red{-3.23\%}& \red{-14.14\%} & \red{-11.93\%} & \red{-7.37\%} &\red{-8.59\%} \\ \hline
(d) \maskmethod $\rightarrow$ No mask          & 0.426                                 & 0.441                                 & 0.349                                 & 0.377                                 & 1.493                                 & 0.828                                 & 0.221                                 & 0.267  \\
~~~~~~Improvement &\red{-7.85\%}  &\red{-6.52\%}  &\red{-2.65\%}  & \red{-1.62\%}&  \red{-6.11\%}&  \red{-7.39\%}& \red{-1.84\%} &\red{-4.30\%} \\ \hline
(e) \maskmethod $\rightarrow$ Random mask      & \textbf{0.394} & \textbf{0.414} & 0.340                                 & 0.372                                 & 1.472                                 & 0.829                                 & 0.223                                 & 0.262  \\
~~~~~~Improvement & 0.25\% & 0.00\% & 0.00\% & \red{-0.27\%}& \red{-4.62\%} & \red{-7.52\%} & \red{-2.76\%} & \red{-2.34\%}\\ \hline
(f) Freeze $f_{\text{vis}}(\cdot)$      &         0.431              &0.428         &   0.358      &  0.380       & 1.442     & 0.773          &   0.246         & 0.288 \\
~~~~~~Improvement &\red{-9.11\%}  & \red{-3.38\%} &\red{-5.29\%}  &\red{-2.43\%} &\red{-2.49\%}  &\red{-0.26\%}  &\red{-13.36\%}  &\red{-12.50\%} \\ \hline
(g) W/o decomposition             & 0.408                                 & 0.424                                 & \textbf{0.338} & 0.373                                 & 1.712                                 & 0.903                                 & 0.219                                 & 0.268 \\
~~~~~~Improvement &\red{-3.29\%}  &\red{-2.42\%}  &0.59\%  &\red{-0.54\%} & \red{-21.68\%} &\red{-17.12\%}  &\red{-0.92\%}  & \red{-4.69\%}\\
\bottomrule[1pt]
\end{tabular}
}
\vspace{-0.2cm}
\end{table*}

\subsection{Ablation Analysis}\label{sec.exp.ablation}

% We validate the design choices of \methodb\ using four datasets. For brevity, we defer the results of \methoda\ to Appendix \ref{app. full baseline results}. Table \ref{tab.ablation} presents the ablation analysis. In (a), we switch the numerical forecaster $f_{\text{num}}(\cdot)$ from a linear model to a \texttt{PatchTST}-style \texttt{Transformer} as described in $\S$\ref{sec.methoda}. In (b), we switch the visual forecaster $f_{\text{vis}}(\cdot)$ from \texttt{MAE} to \texttt{SimMIM} \cite{xie2022simmim}. In (c), the gate-based fusion is changed to a regular sum. In (d), our masking strategy \maskmethod\ is removed. Backcasting and forecasting are performed simultaneously with the full look-back window unmasked. In (e), \maskmethod\ is replaced by random masking. In (f), the entire visual forecaster $f_{\text{vis}}(\cdot)$ is frozen instead of fine-tuning the norm layer. In (g), the backcast-residual mechanism is removed, thus no decomposition is performed. In this case, $f_{\text{num}}(\cdot)$ and $f_{\text{vis}}(\cdot)$ are fed with the same input $\mat{x}^{i}$. Their outputs are directly merged using the gate.

We validate the design of \methodb\ through ablation studies on four datasets; \methoda\ results are deferred to Appendix \ref{app. ablation study} for brevity. Table \ref{tab.ablation} summarizes the analysis: (a) replaces the linear model in $f_{\text{num}}(\cdot)$ with a \texttt{PatchTST}-style \texttt{Transformer} (see $\S$\ref{sec.methoda}); (b) swaps \texttt{MAE} with \texttt{SimMIM} \cite{xie2022simmim} as $f_{\text{vis}}(\cdot)$; (c) replaces the gating fusion with a simple sum; (d) removes \maskmethod, performing backcasting and forecasting on the full, unmasked look-back window; (e) substitutes \maskmethod\ with random masking; (f) freezes the entire $f_{\text{vis}}(\cdot)$ instead of fine-tuning norm layers; and (g) removes the backcast-residual mechanism, feeding both $f_{\text{num}}(\cdot)$ and $f_{\text{vis}}(\cdot)$ the same input $\mat{x}^{i}$ and merging their outputs via gating.

Table \ref{tab.ablation} reveals key insights into \methodb's design. In (a), replacing the linear numerical forecaster with a \texttt{Transformer} slightly degrades performance, likely due to the increased difficulty of jointly training \texttt{Transformer} with LVMs. In (b), \texttt{MAE} outperforms \texttt{SimMIM} as $f_{\text{vis}}(\cdot)$, likely due to its \texttt{ViT}-based reconstruction decoder being better suited for pixel-level tasks like LTSF than \texttt{SimMIM}'s linear decoder, while both models share similar encoder architectures. In (c), gate-based fusion outperforms simple summation, highlighting its adaptability to the distinct outputs of $f_{\text{num}}(\cdot)$ and $f_{\text{vis}}(\cdot)$. (d) and (e) underscore the importance of \maskmethod: removing it ({\em i.e.}, (d) ``No mask'') recovers the full look-back window as the backcasted seasonal component, diminishing the trend signal and weakening $f_{\text{num}}(\cdot)$, while ``Random mask'' ({\em i.e.}, (e)) performs slightly worse due to poorer periodic pattern extraction, which leads to many fluctuations. In $\S$\ref{sec.exp.analysis}, we provide visual examples to compare these masking strategies. In (f), fine-tuning only the norm layers significantly improves performance over freezing, confirming the benefit of coordinated learning between forecasters, as described in $\S$\ref{sec.method.training}. Finally, (g) shows that removing the backcast-residual mechanism causes a major performance drop, affirming its role in effective decomposition. Overall, the LVM decoder, fusion strategy, masking method, training approach, and decomposition mechanism are crucial to \methodb's success.

\subsection{Performance Analysis}\label{sec.exp.analysis}

In this section, we perform an in-depth analysis of \method\ using the same four datasets as in $\S$\ref{sec.exp.ablation}.

\begin{figure*}[!t]
\centering
\includegraphics[width=1.0\linewidth]{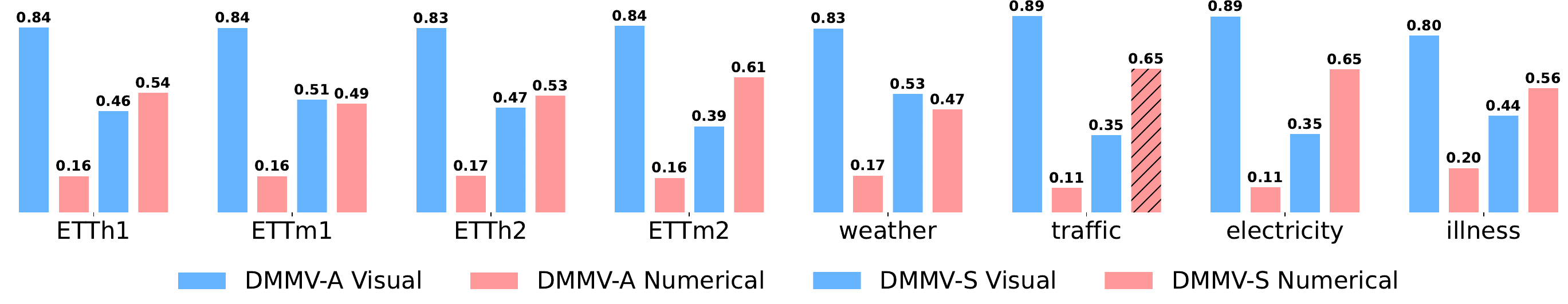}
\caption{Comparing \methoda\ and \methodb\ {\em w.r.t.} gate weights on visual and numerical forecasters.}\label{fig.Gate_Weights}
\vspace{-0.3cm}
\end{figure*}

\textbf{The Difference between \methoda\ and \methodb.} %From Fig. \ref{fig.cdrank}, we observe the superiority of \methodb\ over \methoda. The key difference lies in their decomposition mechanism. One advantage of our gate-based fusion is it enables explaination of the decomposed components. Fig. \ref{fig.Gate_Weights} summarizes the gate weights of the numerical forecaster $f_{\text{num}}(\cdot)$ and visual forecaster $f_{\text{vis}}(\cdot)$ on all datasets (averaged over different prediction lengths). From Fig. \ref{fig.Gate_Weights}, we observe \methodb\ tends to weigh $f_{\text{vis}}(\cdot)$ more than $f_{\text{num}}(\cdot)$, while \methoda\ mostly weigh them equally but is inclined to $f_{\text{num}}(\cdot)$. For \methodb, the weights are fully determined by the forecasting objective, suggesting the importance of $f_{\text{vis}}(\cdot)$. It is noteworthy that although $f_{\text{num}}(\cdot)$ gains a relatively smaller weight, it is indispensable, as validated by the superiority of \methodb\ over the visual-only baseline \texttt{VisionTS} in Table \ref{tab.benchmark}. In contrast, \methoda's weights are partially constrained by the rigid move-average decomposition, which determines how much forecasting power each of $f_{\text{num}}(\cdot)$ and $f_{\text{vis}}(\cdot)$ can contribute. This decision-making is non-adaptive, thus is suboptimal.
Fig. \ref{fig.cdrank} highlights \methodb's superiority over \methoda, largely due to its adaptive decomposition mechanism. A key advantage of the gate-based fusion is its interpretability. As shown in Fig. \ref{fig.Gate_Weights}, which presents average gate weights across datasets, \methodb\ consistently places more weight on $f_{\text{vis}}(\cdot)$, while \methoda\ tends to balance both $f_{\text{num}}(\cdot)$ and $f_{\text{vis}}(\cdot)$ but leans toward $f_{\text{num}}(\cdot)$. In \methodb, these weights are learned based on forecasting performance, emphasizing $f_{\text{vis}}(\cdot)$'s importance. Notably, although $f_{\text{num}}(\cdot)$ receives less weight, it remains essential -- as evidenced by \methodb\ outperforming the visual-only baseline \texttt{VisionTS} in Table \ref{tab.benchmark}. In contrast, \methoda's weights are limited by its fixed moving-average decomposition, leading to a non-adaptive and suboptimal allocation of forecasting roles.

\begin{figure*}[!t]
\centering
\includegraphics[width=0.9\linewidth]{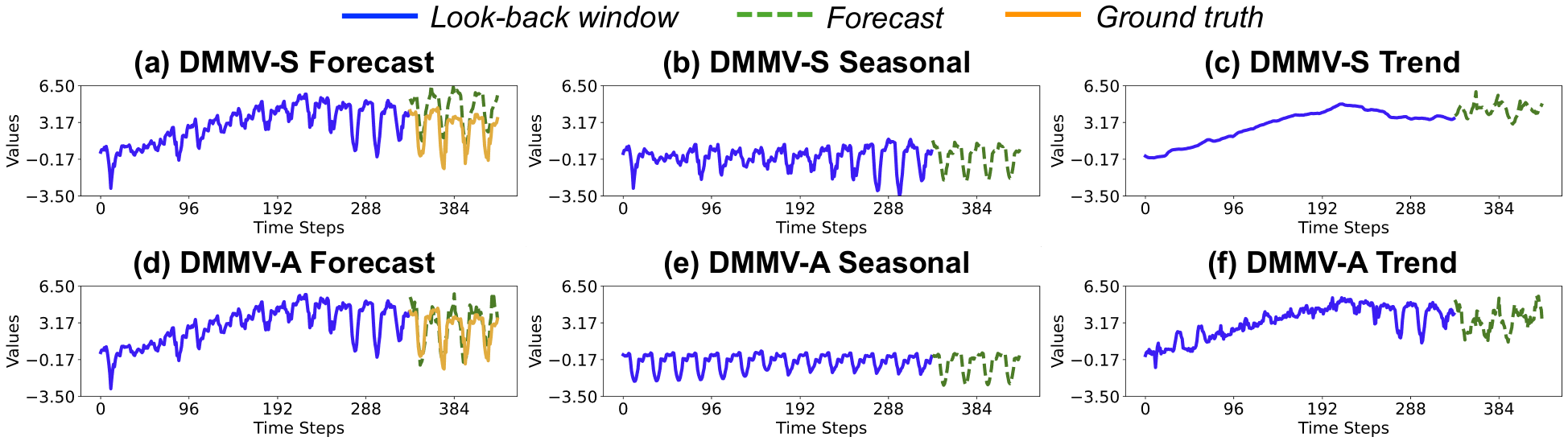}
\caption{The decompositions of \methoda\ and \methodb\ on the same example in ETTh1: (a)(d) input time series and forecasts, (b)(e) seasonal component, and (c)(f) trend component.}\label{fig.decomposition_example}
\vspace{-0.1cm}
\end{figure*}

% We provide some visual examples of \methoda\ and \methodb\ in Fig. \ref{fig.decomposition_example}. More examples can be found in Appendix \ref{app. decomposition visualization}. From Fig. \ref{fig.decomposition_example}, we can see the different decomposition patterns. For \methodb, the seasonal component is smooth and clearly periodic, as we expected. The trend component is also as expected but incorporates some noisy fluctuations. In contrast, \methoda's trend component is smooth. This is because the moving-average smooths out fluctuations in the trend component. As a result, its seasonal component incorporates the fluctuations. This poses a challenge to $f_{\text{vis}}(\cdot)$, which, however, may be more vulnerable to fluctuations than $f_{\text{num}}(\cdot)$, leading to less contributions from $f_{\text{vis}}(\cdot)$, as reflected by its smaller weights in Fig. \ref{fig.Gate_Weights}. Moreover, recent works \cite{lin2024cyclenet} identify the key role of periodic patterns in long-term forecasting. In Fig. \ref{fig.decomposition_example}, we observe a better match between the forecast and the ground truth by \methodb\ than \methoda, which attributes \methodb's better period decomposition.

Fig. \ref{fig.decomposition_example} provides example decompositions by \methoda\ and \methodb\ (additional cases in Appendix \ref{app. decomposition visualization}). \methodb\ produces a smooth, clearly periodic component -- consistent with expectations, and a trend component with some noises. In contrast, \methoda's moving-average yields a smoother trend by absorbing fluctuations, pushing noise into the seasonal component. This makes forecasting harder for $f_{\text{vis}}(\cdot)$, which is more sensitive to fluctuations than $f_{\text{num}}(\cdot)$, resulting in lower weights of $f_{\text{vis}}(\cdot)$ in Fig. \ref{fig.Gate_Weights}. Since periodic patterns are crucial for long-term forecasting, as identified by \cite{lin2024sparsetsf,lin2024cyclenet}, the clearer period separation in \methodb\ leads to forecasts that better match the ground truth.

\begin{figure*}[!t]
\centering
\includegraphics[width=0.8\linewidth]{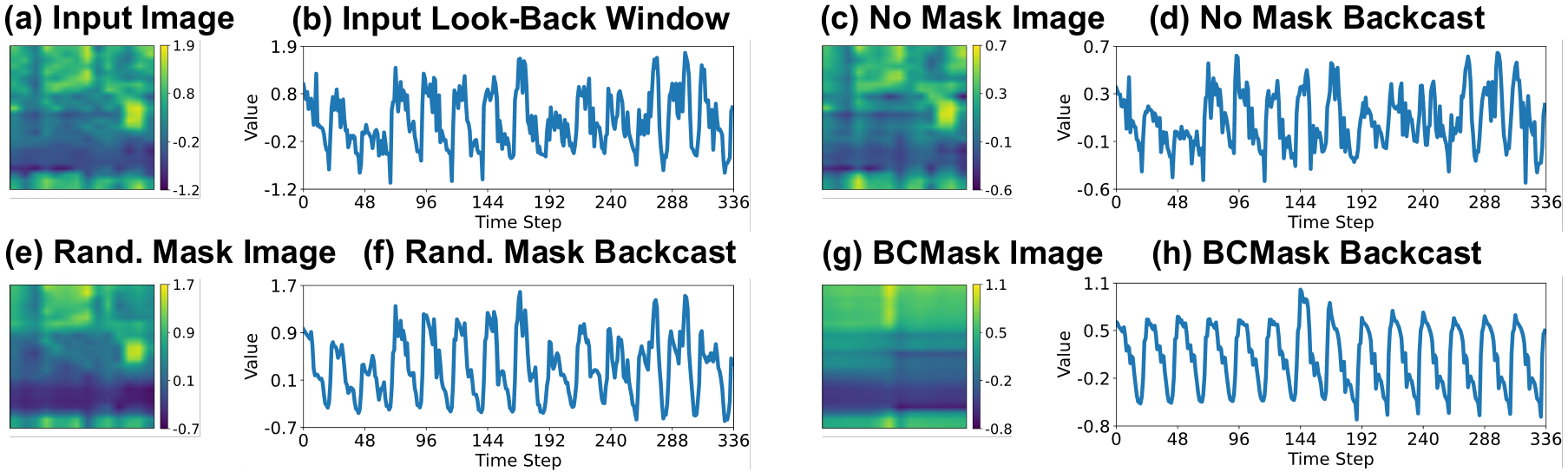}
\caption{Comparison of different masking methods on the same example in ETTh1. (a) image of input look-back window; (c)(e)(g) are images of backcast output by \methodb: (c) uses ``No mask''; (e) uses ``Random mask''; (g) uses \maskmethod. (b)(d)(f)(h) are their recovered time series, respectively.}\label{fig.mask_example}
\vspace{-0.1cm}
\end{figure*}

\textbf{The Effectiveness of \maskmethod.} %Fig. \ref{fig.mask_example} visualizes the images and time series of the backcasts by \methodb\ using \maskmethod\ and its ablations ``No mask'' and ``Random mask'' (see Table \ref{tab.ablation}) on one example. More examples are deferred to Appendix \ref{app. mask visualization}. From Fig. \ref{fig.mask_example}, \maskmethod\ yields an image that is quite smooth along the $x$-axis (temporally ordered segments), recovering a component that successfully extracts clean periodic patterns from the input time series. In contrast, ``No mask'' yields an image that highly resembles the input, delivering no decomposition at all. %On the other hand,
%``Random mask'' yields an image that resembles \maskmethod, demonstrating effectiveness to some extent. Whereas, the patterns are less smooth over the $x$-axis than that of \maskmethod, indicating a suboptimal decomposition.
Fig. \ref{fig.mask_example} compares the backcast results of \methodb\ using \maskmethod, ``No mask'', and ``Random mask'' (as in Table \ref{tab.ablation}) on a sample case; more examples are in Appendix \ref{app. mask visualization}. \maskmethod\ produces a smooth image along the temporal ($x$-axis) segments, effectively capturing clean periodic patterns. In contrast, ``No mask'' closely replicates the input, offering no meaningful decomposition. ``Random mask'' performs moderately well, resembling \maskmethod\ but with less temporal smoothness, indicating a less optimal decomposition.

\begin{figure*}[!t]
\centering
\includegraphics[width=0.85\linewidth]{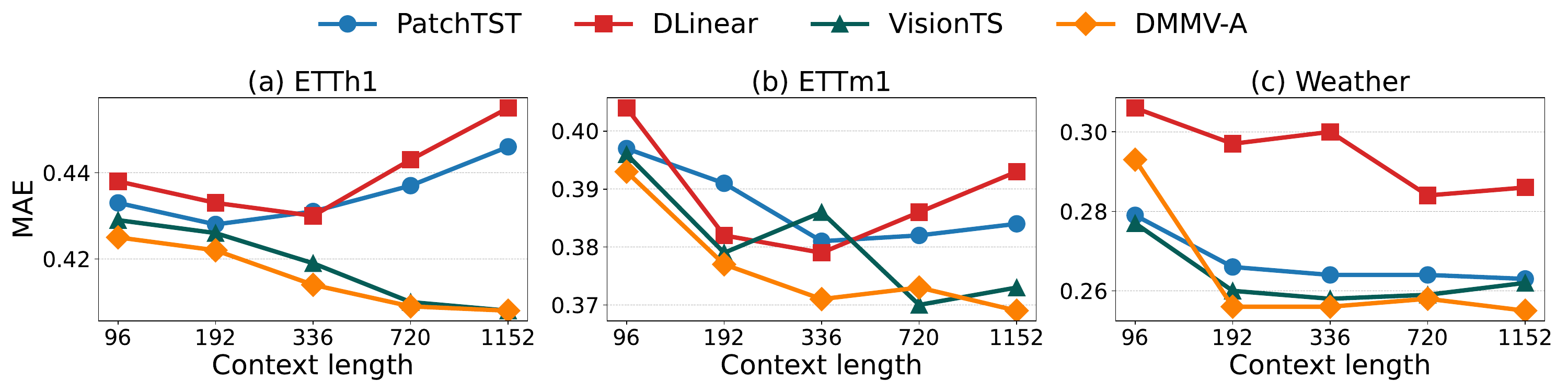}
% \vspace{-1em}
\caption{Average MAE comparison with varying look-back window (or context) lengths.% The MAE is averaged over different prediction lengths.
}\label{fig.context}
\vspace{-0.1cm}
\end{figure*}

% \textbf{Impact of Look-Back Window Length.} %Following \cite{nie2023time,lin2024cyclenet,chen2025visionts}, we assess the impact of look-back window ({\em a.k.a.}, context) length. Fig. \ref{fig.context} compares \methodb\ with two numerical forecasters \texttt{PatchTST}, \texttt{DLinear}, and a visual forecaster \texttt{VisionTS}, which are mostly the single-view ablations of \methodb. The Illness dataset is excluded due to its short time series (966 time steps in total). Fig. \ref{fig.context} uses the MAE metric. MSE comparison is in Appendix \ref{}. From Fig. \ref{fig.context}, \methodb\ and \texttt{VisionTS} are effective in exerting long look-back windows. \texttt{PatchTST} and \texttt{DLinear} show degradations to different extents when the context length is above 336. Moreover, \methodb\ may be more adept at grasping long-term information than \texttt{VisionTS} as indicated by its superiority at context length 1152. This is an advantage of explicit modeling of global trends by \methoda.
%We evaluate the impact of look-back window (context) length.
\textbf{Impact of Look-Back Window.} Fig. \ref{fig.context} compares \methodb\ with a visual forecaster (\texttt{VisionTS}) and two numerical forecasters (\texttt{PatchTST}, \texttt{DLinear}), which can serve as its single-view ablations. %of \methodb.
Illness dataset is excluded due to its short time series (966 time steps). Using MAE metric (MSE results in Appendix \ref{app. Look back windows}), we observe that \methodb\ and \texttt{VisionTS} benefit from longer look-back windows, while \texttt{PatchTST} and \texttt{DLinear} degrade beyond a length of 336. Notably, \methodb\ outperforms \texttt{VisionTS} at length 1152, highlighting the advantage of explicitly modeling global trends.

\section{Conclusion}

This paper introduces \method, a novel MMV framework that leverages LVMs and adaptive decomposition to enhance LTSF. By addressing the inductive bias of LVMs toward periodicity through a tailored backcast-residual decomposition, \method\ effectively integrates numerical and visual perspectives. Extensive experiments on benchmark datasets demonstrate that \method\ outperforms both single-view and SOTA multi-modal baselines, validating its effectiveness. %robustness and generalization capability.
This work highlights the potential of MMVs and LVMs in advancing LTSF, offering a new direction for future research in this domain.% Additionally, we defer our discussion of limitation and broader impact to Appendix \ref{app. limitation and broader impact}.

% \begin{ack}
% acknowledgment
% \end{ack}

\clearpage
\bibliographystyle{abbrv}
\bibliography{ref}

\clearpage
\appendix

\section{Limitation and Broader Impact}\label{app.discussion}

\subsection{Discussion of Limitations}\label{app.discussion.limitation}
% \textbf{Discussion of Limitations}.

Adapting LVMs to LTSF is an emerging area of active research. This work serves a pioneering effort in investigating LVMs within an MMV framework for LTSF. As an initial exploration, we acknowledge some limitations in this work. %First, in Table \ref{tab.benchmark}, the results of \texttt{Time-VLM} and \texttt{CycleNet} on Illness dataset are marked by ``--'' since their papers didn't report the results. By searching them, we found \texttt{Time-VLM}'s code was unavailable at the time of this experiment, while \texttt{CycleNet}'s code was available. Thus we run \texttt{CycleNet} following the standard evaluation protocols \cite{wu2021autoformer} as used in $\S$\ref{sec.exp} to make up its results on Illness dataset and report them in Table \ref{tab. full baseline results} (Appendix \ref{app. full baseline results}).
First, one limitation of the current best LVM forecaster ({\em e.g.}, \texttt{VisionTS}) is its sensitivity to segment length used in image construction due to its inductive bias, as discussed in $\S$\ref{sec.method} (Fig. \ref{fig.bias}). By incorporating an additional numerical view for modeling the global trend, the proposed \method\ is expected to alleviate this sensitivity. In our further analysis in Appendix \ref{app. analysis of inductive bias}, we observe \methodb\ is less sensitive to the change of segment lengths than \texttt{VisionTS} on some datasets, but cannot consistently enhance the robustness over different datasets, despite the improved overall forecasting performance. This may be caused by the higher weights automatically allocated to $f_{\text{vis}}(\cdot)$ than $f_{\text{num}}(\cdot)$ by the gate fusion mechanism (Fig. \ref{fig.Gate_Weights}), which could make the model prone to inherit the behavior of the LVM used in $f_{\text{vis}}(\cdot)$ to some extent, including its sensitivity to segment length, but with a less extent than a sole LVM. As such, a future work to improve \method\ is to reduce such sensitivity to an unnoticeable effect. Second, under our proposed \maskmethod\ strategy, the vision backbone in $f_{\text{vis}}(\cdot)$ is reused three times during training and inference -- once for forecasting and twice for reconstructing different masked parts of the look-back window. The triple use of $f_{\text{vis}}(\cdot)$ could lead to a non-trivial computational overhead. In this work, considering the remarkable performance improvement of \maskmethod\ over other masking strategies (Table \ref{tab.ablation}) and its possibly minimum use of $f_{\text{vis}}(\cdot)$, as analyzed in $\S$\ref{sec.method.adaptive}, we take it as the current solution. However, as a future work, we expect to further reduce the use of $f_{\text{vis}}(\cdot)$ and improve the efficiency, by methods such as joint backcasting and forecasting within a single forward pass or amortizing the use of LVMs across multiple time series samples. Finally, the proposed method shares a similar limitation of the existing LVM forecasters. The imaging process transforms an input time series into a 2D image-like representation to fit pre-trained LVMs, which typically expect a high-resolution input of size such as $224 \times 224\times 3$. Therefore, upsampling is performed on a smaller image obtained from the patched time series %(e.g., of size $24 \times 14 \times 3$)
to fit the input requirements of LVMs. While critical to compatibility, resizing may introduce subtle changes that may distort the original temporal structures to some extent. Thus an imaging process that can reflect the temporal patterns more accurately is in demand in future works.

% \label{app. broader impact}

\subsection{Broader Impact}\label{app.discussion.impact}
% \textbf{Broader Impact}.

LTSF plays a vital role across various domains, including geoscience \cite{ardid2025ergodic}, neuroscience \cite{caro2024brainlm}, energy \cite{koprinska2018convolutional}, healthcare \cite{morid2023time}, and smart city \cite{ma2017learning}. This work proposes a novel MMV framework \method\ that integrates LVMs and a numerical forecaster, which could serve as a groundwork in the emerging area of LVM-based time series analysis and shed some lights on broader areas that integrate LLMs, VLMs, and large multi-modal models (LMMs) for future research on multi-modal and agentic time series analysis. This work does not involve sensitive data, legal risks, or ethical concerns. To the best of our knowledge, it does not adversely affect any specific population. The proposed method could serve as a general-purpose time series forecasting technique with a relatively broad applicability and social acceptability.

\section{Benchmark and Baseline}

\subsection{Benchmark Datasets}\label{app. benchmark datasets}

Following \cite{zhou2021informer,wu2021autoformer,nie2023time,zeng2023transformers,tan2024language,chen2025visionts}, our experiments are conducted on 8 widely used LTSF benchmark datasets that cover a wide range of sampling frequencies, number of variates, levels of periodicity, and real-world domains. The four ETT datasets (ETTh1, ETTh2, ETTm1, ETTm2) record oil temperature from two electric transformers, sampled at 15-minute and hourly intervals. The Weather dataset collects measurements of meteorological indicators in Germany every 10 minutes. The Illness dataset keeps weekly counts of patients and the influenza-like illness ratio from the United States. The Traffic dataset measures hourly road occupancy rates from sensors on San Francisco freeways. The Electricity dataset records hourly electricity consumption of Portuguese clients. Table~\ref{tab. dataset details} summarizes the statistics of the datasets.

% To comprehensively evaluate the performance of our model, we selected eight representative time series benchmark datasets, covering a wide range of sampling frequencies, feature dimensions, and levels of periodicity. These datasets span various real-world domains, including electricity load, weather monitoring, traffic flow, and epidemic trends, each exhibiting diverse temporal patterns. The differences in periodicity and trend characteristics among the datasets pose significant challenges to the model’s generalization ability and robustness across various types of time series. The statistical details of these datasets are summarized in Table~\ref{tab. dataset details}.

\begin{table}[htbp]
\centering
\caption{Statistics of the benchmark datasets. ``Dataset Size'' is organized in (Train, Validation, Test).}
\label{tab. dataset details}

\resizebox{0.85\textwidth}{!}{
%\begin{tabular}{l|cccccccc}
%\toprule[1pt]
%Datasets      & ETTh1 & ETTh2 & ETTm1 & ETTm2 & Illness & Electricity & Weather & Traffic \\\midrule
%\# Variates       & 7     & 7     & 7     & 7     & 7       & 321         & 21      & 862     \\
%Timesteps     & 17420 & 17420 & 69680 & 69680 & 966     & 26304       & 52696   & 17544   \\
%Frequency   & 1h    & 1h    & 15min & 15min & 7d      & 1h          & 10min   & 1h      \\
%Period length & 24    & 24    & 96    & 96    & 54      & 24          & 144     & 24     \\
%\bottomrule[1pt]
%\end{tabular}
\begin{tabular}{lrrcc} \toprule[1pt]
\textbf{Dataset} & \textbf{\# Variates} & \textbf{Series Length} & \textbf{Dataset Size}          & \textbf{Frequency} \\ \midrule
ETTh1       & 7       & 17420         & (8545, 2881, 2881)    & Hourly      \\
ETTh2       & 7       & 17420         & (8545, 2881, 2881)    & Hourly      \\
ETTm1       & 7       & 69680         & (34465, 11521, 11521) & 15 mins     \\
ETTm2       & 7       & 69680         & (34465, 11521, 11521) & 15 mins     \\
Weather     & 321     & 52696         & (36792, 5271, 10540)  & 10 mins     \\
Illness     & 7       & 966           & (617, 74, 170)        & Weekly      \\
Traffic     & 862     & 17544         & (12185, 1757, 3509)   & Hourly      \\
Electricity & 21      & 26304         & (18317, 2633, 5261)   & Hourly     \\ \bottomrule[1pt]
\end{tabular}
}
\end{table}

\subsection{Baselines}
\label{app. baseline}
In the following, we provide a brief description for each baseline method involved in our experiments.

\begin{itemize}
  \item \texttt{Time-VLM} \cite{zhong2025time} integrates time series data with visual views and contextual texts using a pre-trained VLM, \texttt{ViLT}, to enhance forecasting performance.
  \item \texttt{VisionTS} \cite{chen2025visionts} reformulates time series forecasting as an image reconstruction problem using an LVM, \texttt{MAE}, for zero/few/full-shot forecasting.
  \item \texttt{Time-LLM} \cite{jin2023time} reprograms LLMs by aligning time series patches with text tokens, enabling time series forecasting without re-training LLMs.
  \item \texttt{GPT4TS} \cite{zhou2023one} demonstrates that frozen pretrained LLMs, {\em e.g.},  \texttt{GPT}, can be directly applied to a variety of time series tasks with strong performance.
  \item \texttt{CALF} \cite{liu2025calf} adapts LLMs to time series forecasting via cross-modal fine-tuning, bridging the distribution gap between textual and temporal data.
  \item \texttt{CycleNet} \cite{lin2024cyclenet} enhances LTSF by explicitly modeling the periodic patterns in time series through a residual cycle forecasting technique.
  \item \texttt{PatchTST} \cite{nie2023time} introduces a patching strategy and a channel-independence strategy for LTSF. It uses patches of time series as the input to a \texttt{Transformer} to capture the temporal dependency of semantically meaningful tokens ({\em i.e.}, patches).
  \item \texttt{TimesNet} \cite{wu2023timesnet} transforms an input time series into a 2D image-like representation and models temporal variations in the image using inception-like blocks for time series analysis.
  \item \texttt{DLinear} \cite{zeng2023transformers} decomposes an input time series into trend and seasonal components, each of which is modeled by linear layers for time series forecasting.
  \item \texttt{FEDformer} \cite{zhou2022fedformer} incorporates frequency-enhanced attention mechanisms by combining Fourier transforms with seasonal-trend decomposition in a \texttt{Transformer} framework.
  \item \texttt{Autoformer} \cite{wu2021autoformer} introduces an auto-correlation mechanism within a \texttt{Transformer} architecture to capture long-term dependencies in time series data.
  \item \texttt{Stationary} \cite{liu2022non} combines series stationarization and de-stationary attention mechanisms to solve the over-stationarization problem in time series forecasting.%introduces stationarization and de-stationary attention mechanisms in time series data.
  \item \texttt{ETSformer} \cite{woo2022etsformer} decomposes an input time series into interpretable components with exponential smoothing attention and frequency attention for time series forecasting.%combines exponential smoothing techniques with \texttt{Transformer} models, introducing novel attention mechanisms for time series forecasting.
  \item \texttt{Informer} \cite{zhou2021informer} proposes a ProbSparse self-attention mechanism to reduce the computational complexity of LTSF with \texttt{Transformer} models.
\end{itemize}

\section{Implementation Details}\label{app. experiment details}

\begin{algorithm}[!h]
% \small
\DontPrintSemicolon
\SetNoFillComment
% \KwIn{MTS input $\mat{X}=[\mat{x}^{1}, ..., \mat{x}^{D}]^{\top}\in\mathbb{R}^{D\times T}$, ground truth $\mat{Y}\in\mathbb{R}^{D\times H}$}
\KwIn{training dataset $\mathcal{D}_{\text{train}}=\{\mat{X}_{i}, \mat{Y}_{i}\}_{i=1}^{n}$, where $\mat{X}_{i}\in\mathbb{R}^{D\times T}$ is an MTS, $\mat{Y}_{i}\in\mathbb{R}^{D\times H}$ is the ground truth of forecast}

\KwOut{model parameters of \methodb}

% \tcc{Initialize parameters and prepare for training}

\BlankLine

Load pre-trained $f_{\text{vis}}(\cdot)$ and freeze its weights\;

Randomly initialize $f_{\text{num}}(\cdot)$ and the gating parameter $g$\;

\BlankLine

\tcc{stage 1: numerical forecaster training}

% $lr \leftarrow 0.01$, $i \leftarrow 0$\tcp*{Learning rate and iteration counter}

% $b_v \leftarrow \infty$, $p_c \leftarrow 0$\tcp*{Best validation loss and patience counter}

\For{$i\leftarrow 1$ to MaxEpoch}{
% \tcp{Channel Independence}
\For{$(\mat{X},\mat{Y})$ in $\text{Dataloader}(\mathcal{D}_{\text{train}})$}{
\tcc{channel-independence strategy is applied in the following}
    $\mat{\hat{X}}_{\text{season}}, \mat{\hat{Y}}_{\text{season}} \leftarrow f_{\text{vis}}(\mat{X}, \maskmethod)$ \tcp*{backcast/forecast seasonal part}

    $\mat{X_{\text{trend}}} \leftarrow \mat{X} - \mat{\hat{X}}_{\text{season}}$ \tcp*{extract trend component}

    $\mat{\hat{Y}}_{\text{trend}} \leftarrow f_{\text{num}}(\mat{X}_{\text{trend}})$ \tcp*{forecast trend with $f_{\text{num}}(\cdot)$}

    $\mat{\hat{Y}} \leftarrow g \circ \mat{\hat{Y}}_{\text{season}} + (1 - g) \circ \mat{\hat{Y}}_{\text{trend}}$ \tcp*{gate fusion}

    % $mse \leftarrow \text{MSE}(\mat{\hat{y}}, \mat{y})$ \tcp*{Compute loss against ground truth}
    Calculate $\ell_{\text{MSE}}(\mat{\hat{Y}}, \mat{Y})$ \tcp*{calculate MSE loss as specified in $\S$\ref{sec.method.training}}

    % Backpropagate and update parameters of $f_{\text{num}}$ and $g$\;
    Update model parameters of $f_{\text{num}}(\cdot)$ and $g$\;

    % $p_c \leftarrow 0$ and $b_v\leftarrow mse$, if $mse < b_v$, else $p_c \leftarrow p_c + 1$ \tcp*{Early stopping}
    \If{Early stopping condition is TRUE}{
    Break\;
    }

    % $i \leftarrow i + 1$\;
    }

}

\BlankLine

\tcc{stage 2: joint training}

Unfreeze the norm layers in $f_{\text{vis}}(\cdot)$\;

% $lr\leftarrow0.005$, $i\leftarrow 0$ 

% $b_v\leftarrow \infty$, $p_c\leftarrow0$  \tcp*{reset the early stopping controller}

% \While{$i\le 5$ and  $p_c \le 2$}
\For{$i\leftarrow 1$ to MaxEpoch}{
\For{$(\mat{X},\mat{Y})$ in $\text{Dataloader}(\mathcal{D}_{\text{train}})$}{
% \For{\mat{x},\mat{y} in \mat{X},\mat{Y}}{
% \tcp{The computation follows the same procedure as in Stage 1.}
\tcc{Repeat lines 5-9}
% mse$\leftarrow$ callculate MSE(\mat{x}, \mat{y}, $f_{vis}(\cdot)$, $f_{num}(\cdot)$, g) 

%     $mse \leftarrow \text{MSE}(\mat{\hat{y}}, \mat{y})$ 

    % Backpropagate and update parameters of $f_{\text{vis}}$,  $f_{\text{num}}$ and $g$\;

    Update model parameters of $f_{\text{num}}(\cdot)$, norm layers in $f_{\text{vis}}(\cdot)$, and parameter $g$\;

    % $p_c \leftarrow 0$ and $b_v\leftarrow mse$, if $mse < b_v$, else $p_c \leftarrow p_c + 1$ \tcp*{Early stopping}
    \If{Early stopping condition is TRUE}{
    Break\;
    }

    % $i \leftarrow i + 1$\;

}

}

\BlankLine
\caption{The Training Algorithm of \methodb}\label{alg.training}
\end{algorithm}

\subsection{Pre-trained LVM Checkpoints}
\label{app. checkpoint}
As described in $\S$\ref{sec.method.training}, $f_{\text{vis}}(\cdot)$ uses pre-trained LVMs. For \texttt{MAE}, we use the checkpoint released by \textit{Meta Research} \footnote{\url{https://github.com/facebookresearch/mae}}, which was pretrained on $224\times 224 \times 3$ sized images from \textit{ImageNet-1K} \cite{deng2009imagenet} with \texttt{ViT-Base} Backbone. For \texttt{SimMIM}, we adopt the checkpoint released by \textit{Microsoft}\footnote{\url{https://github.com/microsoft/SimMIM}}, which has the same pretraining setting as aforementioned for \texttt{MAE}. For these two LVM backbones, the base versions are adopted to balance the performance and computational costs. 

\subsection{Training Details}
\label{app. training details}
For training the proposed \methoda\ and \methodb\ models, we adopt AdamW optimizer throughout the experiments. The batch size is set to 64 for the ETT datasets and Illness dataset, and set to 8 for the other three datasets to balance training stability and memory consumption.

For both \methoda\ and \methodb, we propose a two-stage training scheme to facilitate effective integration of numerical and visual features:

\begin{itemize}
    \item \textbf{Stage 1} (Numerical forecaster training). In this stage, we freeze all parameters of $f_{\text{vis}}(\cdot)$ and train $f_{\text{num}}(\cdot)$ only. This warm-up step prevents $f_{\text{vis}}(\cdot)$ from updating with unstable gradients caused by the random representations from the under-trained $f_{\text{num}}(\cdot)$. %when cross-modal representations are not yet aligned.
    In this stage, the learning rate is set to 0.01. The training runs up to a maximum of 50 epochs on the training set. Early stopping is applied with a patience of 10 epochs.
    
    \item \textbf{Stage 2} (Joint training). In this stage, we unfreeze the layer normalization parameter in $f_{\text{vis}}(\cdot)$ and jointly train them with $f_{\text{num}}(\cdot)$ to enable deep fusion of visual and numerical views. The learning rate is reduced to 0.005 to preserve learned features and stabilize training. The training at this stage runs up to 5 epochs. Early stopping is applied with a patience of 2 epochs.
\end{itemize}

The detailed training algorithm of \methodb\ is summarized in Algorithm \ref{alg.training}.

\subsection{Running Environment}
\label{app. running enviroment}
The experiments are conducted on a Linux server (kernel 5.15.0-139) with 8x NVIDIA RTX 6000 Ada GPUs (48 GB each). The environment uses Python 3.12.8, PyTorch 2.5.1 with CUDA 12.4 and cuDNN 9.1. The key libraries include NumPy 2.1.3, Pandas 2.2.3, Matplotlib 3.10.0, SciPy 1.15.1, scikit-learn 1.6.1, and torchvision 0.20.1.

\section{More Experimental Results}

\subsection{Comparison with All Baselines}\label{app. full baseline results}

% To provide a comprehensive reference, we present the complete comparison results of \methodb, \methoda, and other commonly used baselines, see

Table \ref{tab. full baseline results} provides the full results of comparing \methodb\ and \methoda\ with all of the 14 baseline methods, which complements Table \ref{tab.benchmark} in the paper. In Table \ref{tab. full baseline results}, \texttt{Time-VLM}'s results on Illness dataset is marked by ``--'' since its paper doesn't report the results and its code is not publicly available at the time of this %submission.
experiment. \texttt{CycleNet}'s paper doesn't report its results on Illness dataset, so we run its code and reproduce its results on Illness dataset in Table \ref{tab. full baseline results}.

From Table \ref{tab. full baseline results}, we can observe that \methodb\ maintains a clear advantage when compared against all of the baseline methods. It achieves 41 first-place results, significantly surpassing the second-best method. Additionally, taking a closer look at all compared methods, MMV-based methods %(e.g., Time-LVM)
LVM-based methods, and decomposition-based methods %(e.g., Cyclenet)
demonstrate superiority over other baseline methods. This suggests %our hypothesis regarding the effectiveness of both
the synergy of MMV framework, LVMs, and decomposition strategy, which are explored by the proposed \method\ model.

\begin{sidewaystable}[p]
\centering
\caption{
Full LTSF performance of the compared methods on the benchmark datasets. Lower MSE and MAE indicate better performance. The best performance is highlighted in \textcolor{red}{\textbf{red}}. \texttt{Time-VLM} results on the Illness dataset are unavailable in \cite{zhong2025time}. Its code was not publicly available at the time of this %submission.
experiment. As such, its results on Illness dataset are marked by ``--''.
}
\resizebox{\textwidth}{!}{
\label{tab. full baseline results}
\begin{tabular}{c|c|cccc|cccccccccccccccccccccccccccc}
\toprule[1pt]
                              & Model  & \multicolumn{2}{c}{\methodb}                                            & \multicolumn{2}{c|}{\methoda}                                 & \multicolumn{2}{c}{Time-VLM}                                                              & \multicolumn{2}{c}{VisionTS}                                                   & \multicolumn{2}{c}{Time-LLM}                  & \multicolumn{2}{c}{GPT4TS}                                                                & \multicolumn{2}{c}{CALF}                   & \multicolumn{2}{c}{CycleNet}                                                              & \multicolumn{2}{c}{PatchTST}                                                              & \multicolumn{2}{c}{TimesNet} & \multicolumn{2}{c}{DLinear} & \multicolumn{2}{c}{FEDformer} & \multicolumn{2}{c}{Autoformer} & \multicolumn{2}{c}{Stationary} & \multicolumn{2}{c}{ETSformer} & \multicolumn{2}{c}{Informer} \\\midrule
                              & Metric & MSE                                         & MAE                                         & MSE                                   & MAE                                   & MSE                                         & MAE                                         & MSE                                         & MAE                                         & MSE   & MAE                                   & MSE                                         & MAE                                         & MSE   & MAE                                & MSE                                         & MAE                                         & MSE                                         & MAE                                         & MSE           & MAE          & MSE          & MAE          & MSE           & MAE           & MSE            & MAE           & MSE            & MAE           & MSE           & MAE           & MSE           & MAE          \\
\end{tabular}
}
\end{sidewaystable}

\subsection{Further Analysis of The Inductive Bias}\label{app. analysis of inductive bias}

\begin{table}[!t]
\centering
\caption{Forecasting performance of an LVM {\em w.r.t.} varying segment length on a synthetic dataset. The function of the synthetic time series is
\( x(t) = A(t) \cdot \sin\left( \frac{2\pi t}{P} \right) \), where the period $P=24$ and the amplitude function \(A(t)\) decreases linearly over time.} \label{tab. simulate prediction resultes}
\resizebox{0.9\textwidth}{!}{
\begin{tabular}{c|ccccccccc}
\toprule[1pt]
Segment   Length & 16    & 20    & \textbf{24}    & 28    & 32    & 36    & 40    & 44    & \textbf{48}    \\ \midrule
MSE              & 0.043 & 0.099 & 0.001 & 0.147 & 0.154 & 0.143 & 0.221 & 0.114 & 0.002 \\
MAE              & 0.177 & 0.257 & 0.024 & 0.342 & 0.347 & 0.315 & 0.408 & 0.289 & 0.045\\ \bottomrule[1pt]
\end{tabular}
}
\end{table}

A contribution of our work lies in the in-depth analysis of an inductive bias of the current best LVM forecasters. In $\S$\ref{sec.method}, we have discussed the impact of the alignment of the segment length and the period of time series on model performance. We find that the LVM exhibits a strong \textit{inter-period consistency} when applied to synthetic data. The function of the synthetic time series is \( x(t) = A(t) \cdot \sin\left( \frac{2\pi t}{P} \right) \), where the period $P$ is set to 24 and the amplitude function \(A(t)\) decreases linearly over time. The forecasts are more accurate when the segment length is a multiple of the period ({\em e.g.}, 24, 48) than other values. This section provides detailed quantitative results on the synthetic data in Table \ref{tab. simulate prediction resultes}. From Table \ref{tab. simulate prediction resultes}, the fluctuations in MSEs and MAEs across different segment lengths other than 24 and 48 support the findings of the inductive bias toward ``forecasting periods''.

\begin{figure*}[!t]
\centering
\includegraphics[width=0.99\linewidth]{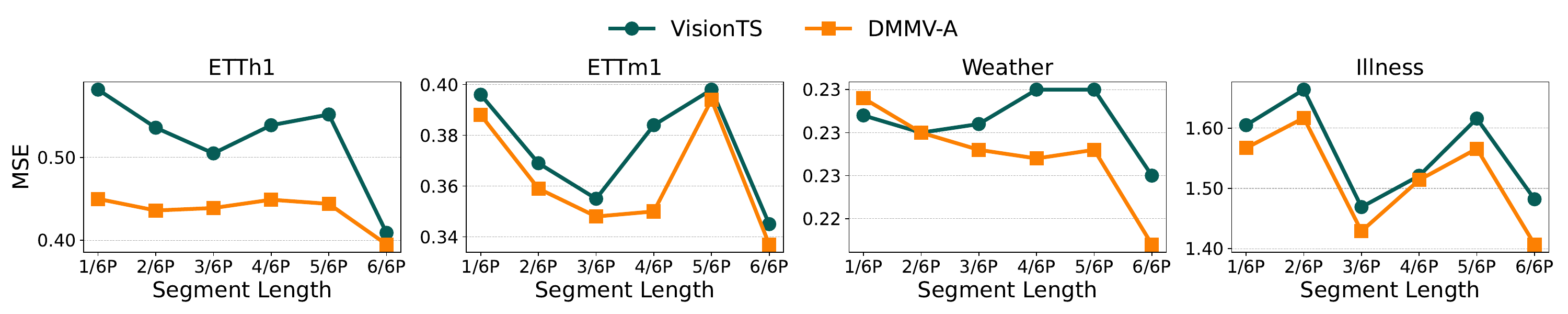}
% \vspace{-1em}
\caption{MSE Performance of \methodb\ and \texttt{VisionTS} {\em w.r.t.} varying segment length that is used in image construction. The $x$-axis indicates the segment length varies from $\frac{1}{6}$ period to $\frac{6}{6}$ period.}\label{fig. period shift}
% \vspace{-0.3cm}
\end{figure*}

In addition, we evaluate the performance of the proposed method \methodb\ and \texttt{VisionTS} {\em w.r.t.} varying segment lengths to compare their robustness to the change of segment length. Fig. \ref{fig. period shift} summarizes the results in terms of MSE on four benchmark datsets, where the segment length varies from $\frac{P}{6}$ to $\frac{6P}{6}$ and $P$ is a period of the input time series. From Fig. \ref{fig. period shift}, we have several observations. First, \methodb\ consistently outperforms \texttt{VisionTS}, validating the effectiveness of the proposed MMV framework. Second, in contrast to \texttt{VisionTS}, \methodb\ exhibits a better robustness to the change of segment length on ETTh1 and Weather datasets, but has a similar sensitivity to the change of segment length as \texttt{VisionTS} on ETTm1 and Illness datasets. This implies that by incorporating $f_{\text{num}}(\cdot)$, \methodb\ can alleviate $f_{\text{vis}}(\cdot)$'s sensitivity to the inductive bias to some extent. However, the current \methodb\ does not fully mitigate this limitation, suggesting a future work for method development as discussed in Appendix \ref{app.discussion}.
% It may be caused by the higher weights that are automatically allocated to $f_{\text{vis}}(\cdot)$ than $f_{\text{num}}(\cdot)$ by the gate fusion mechanism (Fig. \ref{fig.Gate_Weights}), which could make the model prone to inherit the behavior of the LVM used in $f_{\text{vis}}(\cdot)$ to some extent, including its sensitivity to segment length, but with a less extent than a sole LVM. As such, a future work to improve \method\ is to reduce such sensitivity to an unnoticeable effect.
% on certain datasets such as ETTh1, our method demonstrates less sensitivity to inductive bias, yielding more stable and reliable forecasts.

\subsection{Ablation Study}
\label{app. ablation study}
In Table \ref{tab.ablation} ($\S$\ref{sec.exp.ablation}), we provide ablation analyses for \methodb. %To further support our conclusions, we provide more detailed results in this section. Specifically,
Table \ref{tab.ablation of methoda} provides the ablation analysis for \methoda, where MSE and MAE are averaged over different prediction lengths. In addition, Tables \ref{tab.full ablation of methodb} (Table \ref{tab.full ablation of methoda}) includes the full results for Table \ref{tab.ablation} (Table \ref{tab.ablation of methoda}) with all prediction lengths.% and \ref{tab.full ablation of methoda} contain comprehensive numerical results for both \methodb\ and \methoda\ under various settings, including different prediction lengths and datasets, serving as a reference for in-depth analysis and reproducibility.

In Table \ref{tab.ablation of methoda}, from (a), replacing the linear numerical forecaster with \texttt{PatchTST} can slightly improve the performance of \methoda, likely because \methoda\ relies more on the predictions from the numerical view than visual view (Fig. \ref{fig.Gate_Weights}). Therefore, in this case, increasing the complexity of the numerical model can improve the ability of $f_{\text{num}}(\cdot)$ and finally improve the overall performance. From (b), replacing \texttt{MAE} with \texttt{SimMiM} reduces the overall performance, this is the same as the findings in Table \ref{tab.ablation} for \methodb. From (c), gate-based fusion outperforms simple summation for \methoda, highlighting the effectiveness of gate fusion. From (d), fine-tuning the norm layers of $f_{\text{vis}}(\cdot)$ improves the performance for \methoda, suggesting the used fine-tuning strategy.

% The detailed experimental results of \methodb\ and \methoda\ in Table \ref{tab.full ablation of methodb} and Table \ref{tab.full ablation of methoda} further support the conclusions. %the correctness and reliability of our previous findings. By comparing the results across various datasets and prediction lengths, we observe that the relative performance between methods remains largely consistent. This stable trend indicates that the conclusions drawn from the ablation studies are not incidental but possess general applicability across different scenarios.
% Therefore, using the mean values for analysis is reasonable and effective, as it captures the overall performance trends and strengths of the models.

\begin{table*}[!h]
\centering
\caption{Ablation analysis of \methoda. MSE and MAE are averaged over different prediction lengths. Lower MSE and MAE are better. ``Improvement'' of each ablation is relative to \methoda.}\label{tab.ablation of methoda}
\vspace{0.3cm}
\scriptsize
\setlength{\tabcolsep}{5pt}{
\begin{tabular}{l|cc|cc|cc|cc}
\toprule[1pt]
\textbf{Dataset ($\rightarrow$)} & \multicolumn{2}{c|}{ETTh1}                                                     & \multicolumn{2}{c|}{ETTm1}                                                     & \multicolumn{2}{c|}{Illness}                                                   & \multicolumn{2}{c}{Weather} \\ \hline %\cmidrule{1-3} \cmidrule{4-5} \cmidrule{6-7} \cmidrule{8-9}
\textbf{Method ($\downarrow$),~~~Metric ($\rightarrow$)}  & MSE                                   & MAE                                   & MSE                                   & MAE                                   & MSE                                   & MAE                                   & MSE                                   & MAE \\ \midrule
\methoda\                 & 0.405                          & 0.426                          & 0.349                          & 0.382                          & 1.520                           & 0.813                           & 0.244                          & 0.281 \\ \hline
(a) $f_{\text{num}}(\cdot)\rightarrow$ \texttt{Transformer}         & \textbf{0.402}                 & \textbf{0.423}                 & \textbf{0.342}                 & \textbf{0.376}                 & 1.544                           & 0.841                           & \textbf{0.229}                 & \textbf{0.264} \\
~~~~~~Improvement & 0.74\%                         & 0.47\%                         & 2.01\%                         & 1.31\%                         & \red{ -1.65\%}  & \red{ -3.44\%}  & 6.15\%                         & 6.41\%                          \\ \hline
(b) $f_{\text{vis}}(\cdot)\rightarrow$ \texttt{SimMIM}        &  0.415                          & \textbf{0.423}                 & 0.355                          & 0.382                          & 1.810                           & 0.875                           & 0.233                          & 0.272                           \\
~~~~~~Improvement &  \red{ -2.47\%} & 0.47\%                         & \red{ -1.72\%} & 0.00\%                         & \red{ -19.16\%} & \red{ -7.63\%}  & 4.92\%                         & 3.20\%  \\ \hline
(c) Gate $\rightarrow$ Sum        & 0.419                          & 0.435                          & 0.355                          & 0.379                          & \textbf{1.453}                  & \textbf{0.790}                  & 0.256                          & 0.297                          \\
~~~~~~Improvement &\red{ -3.46\%} & \red{ -2.24\%} & \red{ -2.01\%} & 0.52\%                         & 4.34\%                          & 2.95\%                          & \red{ -4.51\%} & \red{ -5.69\%} \\\hline
(d) Freeze $f_{\text{vis}}(\cdot)$      &  0.436                          & 0.442                          & 0.368                          & 0.386                          & 2.125                           & 0.969                           & 0.251                          & 0.288                               \\
~~~~~~Improvement & \red{ -7.41\%} & \red{ -3.76\%} & \red{ -5.75\%} & \red{ -1.05\%} & \red{ -39.83\%} & \red{ -19.07\%} & \red{ -2.46\%} & \red{ -2.14\%} \\ 
\bottomrule[1pt]
\end{tabular}
}
% \vspace{-0.2cm}
\end{table*}

\begin{table}[!h]
\centering
\caption{Full results of the ablation analysis of \methodb. Lower MSE and MAE are better. The Illness dataset uses prediction lengths of $\{24, 36, 48, 60\}$ due to its short time series (in total 966 time steps), which is different from the prediction lengths of other datasets.}\label{tab.full ablation of methodb}
\scriptsize
\begin{tabular}{l|c|cc|cc|cc|cc}
\toprule[1pt]
\multicolumn{2}{l}{\textbf{Dataset($\rightarrow$)}} & \multicolumn{2}{c}{ETTh1} & \multicolumn{2}{c}{ETTm1} & \multicolumn{2}{c}{Illness} & \multicolumn{2}{c}{Weather} \\
\textbf{Method($\downarrow$),~~~Metric($\rightarrow$)}             & \textbf{Length}  & MSE         & MAE         & MSE         & MAE         & MSE          & MAE          & MSE          & MAE          \\ \midrule
\multirow{5}{*}{\methodb}            & 96     & 0.354       & 0.389       & 0.279       & 0.329       & 1.409        & 0.754        & 0.143        & 0.195        \\
                   & 192    & 0.393       & 0.405       & 0.317       & 0.357       & 1.290        & 0.745        & 0.187        & 0.242        \\
                   & 336    & 0.387       & 0.413       & 0.351       & 0.381       & 1.499        & 0.810        & 0.237        & 0.273        \\
                   & 720    & 0.445       & 0.450       & 0.411       & 0.415       & 1.428        & 0.773        & 0.302        & 0.315        \\
                   & Avg.   & 0.395       & 0.414       & 0.340       & 0.371       & 1.407        & 0.771        & 0.217        & 0.256        \\\midrule
\multirow{5}{*}{(a) $f_{\text{num}}(\cdot)\rightarrow\texttt{Transformer}$}        & 96     & 0.357       & 0.389       & 0.279       & 0.329       & 1.604        & 0.823        & 0.145        & 0.193        \\
                   & 192    & 0.407       & 0.420       & 0.318       & 0.359       & 1.250        & 0.742        & 0.187        & 0.239        \\
                   & 336    & 0.389       & 0.411       & 0.352       & 0.382       & 1.555        & 0.803        & 0.241        & 0.283        \\
                   & 720    & 0.474       & 0.462       & 0.407       & 0.416       & 1.359        & 0.774        & 0.301        & 0.326        \\
                   & Avg.   & 0.407       & 0.421       & 0.339       & 0.372       & 1.442        & 0.786        & 0.219        & 0.260        \\\midrule
\multirow{5}{*}{(b) $f_{\text{vis}}(\cdot)\rightarrow\texttt{SimMiM}$}              & 96     & 0.358       & 0.383       & 0.301       & 0.348       & 1.729        & 0.832        & 0.145        & 0.194        \\
                   & 192    & 0.405       & 0.41        & 0.325       & 0.363       & 1.643        & 0.734        & 0.192        & 0.242        \\
                   & 336    & 0.412       & 0.414       & 0.354       & 0.383       & 1.689        & 0.845        & 0.241        & 0.275        \\
                   & 720    & 0.453       & 0.452       & 0.398       & 0.412       & 1.534        & 0.845        & 0.328        & 0.332        \\
                   & Avg.   & 0.407       & 0.415       & 0.345       & 0.377       & 1.649        & 0.814        & 0.227        & 0.261        \\\midrule
\multirow{5}{*}{(c) Gate $\rightarrow$ Sum}                 & 96     & 0.373       & 0.400       & 0.286       & 0.339       & 1.728        & 0.845        & 0.156        & 0.214        \\
                   & 192    & 0.414       & 0.424       & 0.329       & 0.369       & 1.423        & 0.795        & 0.204        & 0.261        \\
                   & 336    & 0.411       & 0.422       & 0.364       & 0.392       & 1.693        & 0.920        & 0.258        & 0.302        \\
                   & 720    & 0.457       & 0.461       & 0.427       & 0.430       & 1.580        & 0.890        & 0.315        & 0.335        \\
                   & Avg.   & 0.414       & 0.427       & 0.352       & 0.383       & 1.606        & 0.863        & 0.233        & 0.278        \\\midrule
\multirow{5}{*}{(d)\maskmethod $\rightarrow$ No mask }               & 96     & 0.384       & 0.402       & 0.288       & 0.342       & 1.628        & 0.840        & 0.145        & 0.198        \\
                   & 192    & 0.413       & 0.440       & 0.325       & 0.363       & 1.325        & 0.796        & 0.191        & 0.244        \\
                   & 336    & 0.434       & 0.448       & 0.361       & 0.384       & 1.606        & 0.865        & 0.241        & 0.285        \\
                   & 720    & 0.474       & 0.473       & 0.421       & 0.419       & 1.414        & 0.811        & 0.308        & 0.340        \\
                   & Avg.   & 0.426       & 0.441       & 0.349       & 0.377       & 1.493        & 0.828        & 0.221        & 0.267        \\\midrule
\multirow{5}{*}{(e)\maskmethod $\rightarrow$ Random mask }           & 96     & 0.348       & 0.384       & 0.279       & 0.329       & 1.618        & 0.859        & 0.146        & 0.197        \\
                   & 192    & 0.388       & 0.405       & 0.318       & 0.360       & 1.318        & 0.798        & 0.189        & 0.240        \\
                   & 336    & 0.383       & 0.404       & 0.350       & 0.381       & 1.560        & 0.858        & 0.243        & 0.282        \\
                   & 720    & 0.458       & 0.462       & 0.414       & 0.418       & 1.392        & 0.800        & 0.312        & 0.328        \\
                   & Avg.   & 0.394       & 0.414       & 0.340       & 0.372       & 1.472        & 0.829        & 0.223        & 0.262        \\\midrule
\multirow{5}{*}{(f) Freeze $f_{\text{vis}}(\cdot)$ }                & 96     & 0.389       & 0.402       & 0.293       & 0.342       & 1.482        & 0.761        & 0.161        & 0.224        \\
                   & 192    & 0.434       & 0.425       & 0.335       & 0.367       & 1.218        & 0.694        & 0.203        & 0.287        \\
                   & 336    & 0.431       & 0.428       & 0.372       & 0.389       & 1.58         & 0.82         & 0.285        & 0.302        \\
                   & 720    & 0.468       & 0.457       & 0.431       & 0.422       & 1.489        & 0.815        & 0.335        & 0.338        \\
                   & Avg.   & 0.431       & 0.428       & 0.358       & 0.380       & 1.442        & 0.773        & 0.246        & 0.288        \\\midrule
\multirow{5}{*}{(g) W/o decomposition}      & 96     & 0.352       & 0.387       & 0.274       & 0.329       & 1.728        & 0.938        & 0.143        & 0.195        \\
                   & 192    & 0.402       & 0.414       & 0.315       & 0.358       & 1.841        & 0.940        & 0.187        & 0.242        \\
                   & 336    & 0.391       & 0.410       & 0.347       & 0.382       & 1.672        & 0.886        & 0.237        & 0.284        \\
                   & 720    & 0.487       & 0.486       & 0.417       & 0.422       & 1.606        & 0.846        & 0.309        & 0.350        \\
                   & Avg.   & 0.408       & 0.424       & 0.338       & 0.373       & 1.712        & 0.903        & 0.219        & 0.268       \\ \bottomrule[1pt]
\end{tabular}
\end{table}

\begin{table}[!h]
\centering
\caption{Full results of the ablation analysis of \methoda. Lower MSE and MAE are better. The Illness dataset uses prediction lengths of $\{24, 36, 48, 60\}$ due to its short time series (in total 966 time steps), which is different from the prediction lengths of other datasets.}\label{tab.full ablation of methoda}
\scriptsize
\begin{tabular}{l|c|cc|cc|cc|cc}
\toprule[1pt]
\multicolumn{2}{l}{\textbf{Dataset($\rightarrow$)}} & \multicolumn{2}{c}{ETTh1} & \multicolumn{2}{c}{ETTm1} & \multicolumn{2}{c}{Illness} & \multicolumn{2}{c}{Weather} \\
\textbf{Method($\downarrow$),~~~Metric($\rightarrow$)}             & \textbf{Length}  & MSE         & MAE         & MSE         & MAE         & MSE          & MAE          & MSE          & MAE          \\ \midrule
\multirow{5}{*}{\methoda}      & 96    & 0.350       & 0.388       & 0.296       & 0.349       & 1.638        & 0.838        & 0.168        & 0.218        \\
                             & 192   & 0.399       & 0.420       & 0.328       & 0.370       & 1.323        & 0.753        & 0.220        & 0.259        \\
                             & 336   & 0.399       & 0.415       & 0.369       & 0.393       & 1.644        & 0.851        & 0.267        & 0.304        \\
                             & 720   & 0.472       & 0.479       & 0.401       & 0.414       & 1.473        & 0.810        & 0.322        & 0.343        \\
                             & Avg.  & 0.405       & 0.426       & 0.349       & 0.382       & 1.520        & 0.813        & 0.244        & 0.281        \\ \midrule
\multirow{5}{*}{(a) $f_{\text{num}}(\cdot)\rightarrow\texttt{Transformer}$}  & 96    & 0.352       & 0.387       & 0.286       & 0.339       & 1.613        & 0.829        & 0.148        & 0.194        \\
                             & 192   & 0.401       & 0.420       & 0.325       & 0.364       & 1.417        & 0.825        & 0.193        & 0.240        \\
                             & 336   & 0.395       & 0.415       & 0.354       & 0.387       & 1.610        & 0.853        & 0.246        & 0.280        \\
                             & 720   & 0.460       & 0.471       & 0.401       & 0.414       & 1.536        & 0.858        & 0.330        & 0.341        \\
                             & Avg.  & 0.402       & 0.423       & 0.342       & 0.376       & 1.544        & 0.841        & 0.229        & 0.264        \\\midrule
\multirow{5}{*}{(b) $f_{\text{vis}}(\cdot)\rightarrow\texttt{SimMiM}$}       & 96    & 0.366       & 0.391       & 0.323       & 0.360       & 1.923        & 0.901        & 0.153        & 0.210        \\
                             & 192   & 0.412       & 0.420       & 0.331       & 0.364       & 1.812        & 0.863        & 0.194        & 0.248        \\
                             & 336   & 0.419       & 0.420       & 0.361       & 0.386       & 1.793        & 0.854        & 0.245        & 0.279        \\
                             & 720   & 0.464       & 0.461       & 0.404       & 0.416       & 1.712        & 0.883        & 0.339        & 0.352        \\
                             & Avg.  & 0.415       & 0.423       & 0.355       & 0.382       & 1.810        & 0.875        & 0.233        & 0.272        \\\midrule
\multirow{5}{*}{(c) Gate $\rightarrow$ Sum}         & 96    & 0.356       & 0.389       & 0.300       & 0.346       & 1.503        & 0.763        & 0.183        & 0.234        \\
                             & 192   & 0.403       & 0.417       & 0.334       & 0.365       & 1.350        & 0.746        & 0.236        & 0.277        \\
                             & 336   & 0.414       & 0.426       & 0.362       & 0.385       & 1.530        & 0.820        & 0.271        & 0.308        \\
                             & 720   & 0.504       & 0.506       & 0.424       & 0.420       & 1.429        & 0.830        & 0.333        & 0.369        \\
                             & Avg.  & 0.419       & 0.435       & 0.355       & 0.379       & 1.453        & 0.790        & 0.256        & 0.297        \\\midrule
\multirow{5}{*}{(d) Freeze $f_{\text{vis}}(\cdot)$ }      & 96    & 0.386       & 0.404       & 0.306       & 0.352       & 1.966        & 0.921        & 0.156        & 0.225        \\
                             & 192   & 0.436       & 0.434       & 0.347       & 0.375       & 2.050        & 0.945        & 0.240        & 0.261        \\
                             & 336   & 0.436       & 0.440       & 0.377       & 0.392       & 2.223        & 0.999        & 0.271        & 0.312        \\
                             & 720   & 0.484       & 0.488       & 0.443       & 0.424       & 2.259        & 1.009        & 0.335        & 0.353        \\
                             & Avg.  & 0.436       & 0.442       & 0.368       & 0.386       & 2.125        & 0.969        & 0.251        & 0.288       \\ \bottomrule[1pt]
\end{tabular}
\end{table}

\subsection{Additional Visualizations on Decomposition}\label{app. decomposition visualization}

\begin{figure*}[!t]
\centering
\includegraphics[width=0.9\linewidth]{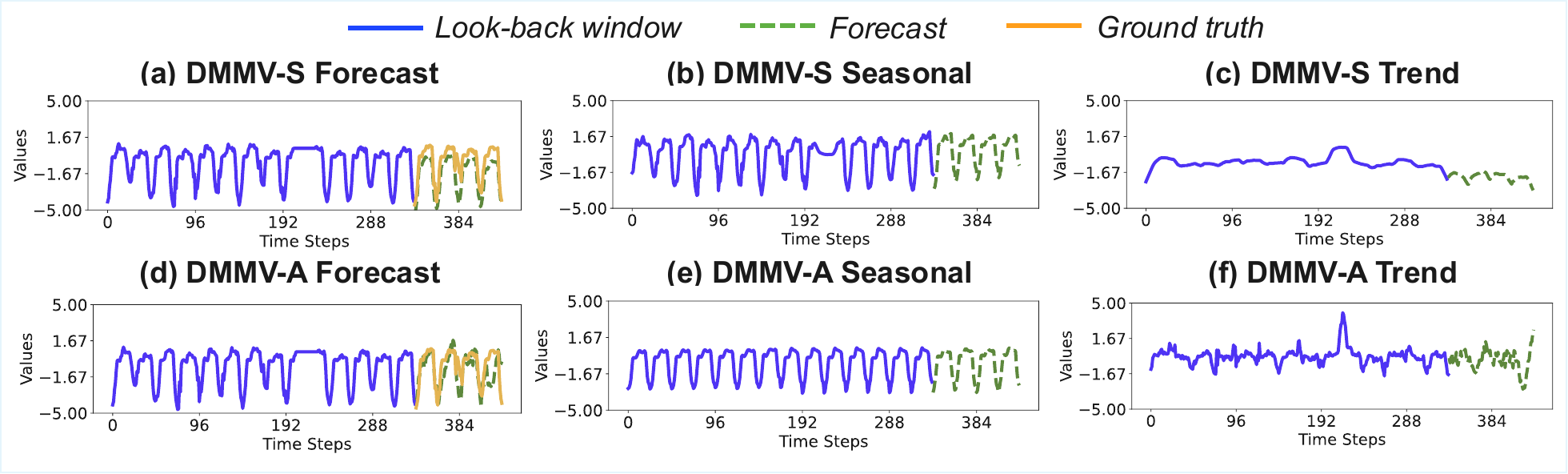}
% \caption{The decompositions of \methoda\ and \methodb\ on the example having anomaly in ETTh1}\label{fig.additional_decompose_example1}
\caption{The decompositions of \methoda\ and \methodb\ on the same example in ETTh1: (a)(d) input time series and forecasts, (b)(e) seasonal component, and (c)(f) trend component.}\label{fig.additional_decompose_example1}
\vspace{-0.2cm}
\end{figure*}

\begin{figure*}[!h]
\centering
\includegraphics[width=0.9\linewidth]{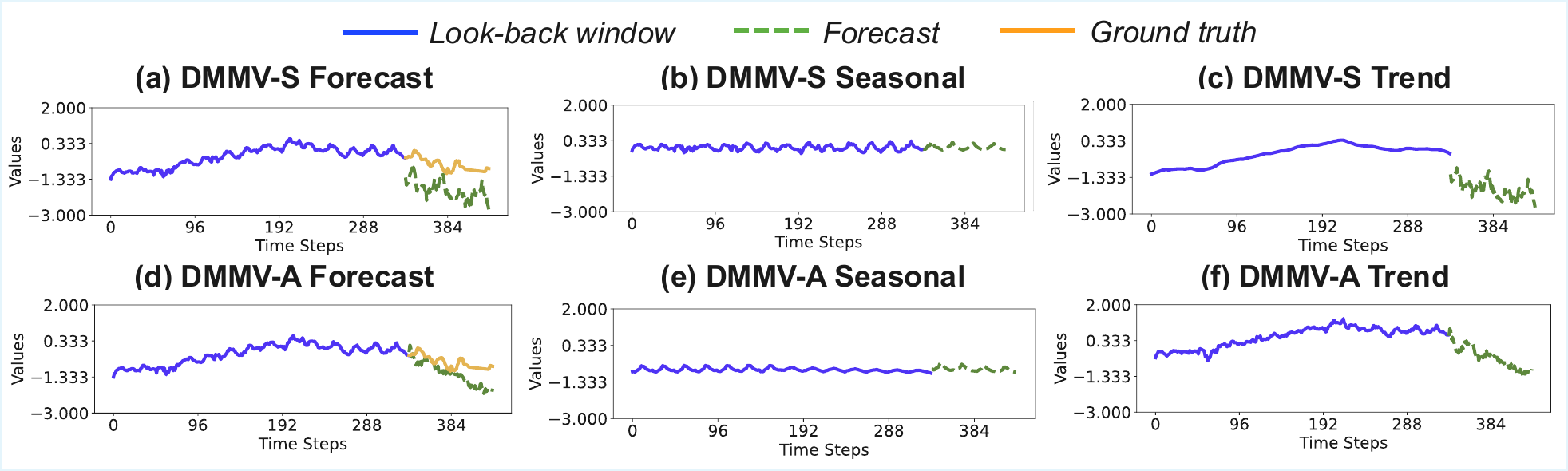}
% \caption{The decompositions of \methoda\ and \methodb\ on the same example having no seasonal in ETTh2}\label{fig.additional_decompose_example2}
\caption{The decompositions of \methoda\ and \methodb\ on the same example in ETTh2: (a)(d) input time series and forecasts, (b)(e) seasonal component, and (c)(f) trend component.}\label{fig.additional_decompose_example2}
\vspace{-0.2cm}
\end{figure*}

Fig.~\ref{fig.additional_decompose_example1} and Fig.~\ref{fig.additional_decompose_example2} several more examples the decomposed time series of \methoda\ and \methodb. %on more challenging types of input series, further validating the robustness and generalization ability of the models.
Fig.~\ref{fig.additional_decompose_example1} illustrates a case where the series has a localized periodic anomaly at time step around 192, which poses a challenge for detecting periodic patterns. In this case, \methodb\ effectively suppresses the influence of the anomaly and extracts a clear periodic pattern from the time series series. In contrast, \methoda\ is affected by the anomaly and fails to capture a smooth periodic pattern. Fig.~\ref{fig.additional_decompose_example2} is an example with weak periodicity, where the periodic signal is either faint or overwhelmed by trend. In this case, \methodb\ is able to extract and utilize the underlying periodicity to produce reasonable forecasts, which is better than \methoda, suggesting the importance of the proposed adaptive decomposition method. In summary, the results demonstrate that \methodb\ has a strong modeling ability of temporal structures and robustness to fluctuations even when dealing with anomalous or weakly periodic time series, validating its reliability and applicability across a broad range of scenarios.

\subsection{Additional Visualizations on Masking Strategies}
\label{app. mask visualization}

\begin{figure*}[!h]
\centering
\includegraphics[width=0.75\linewidth]{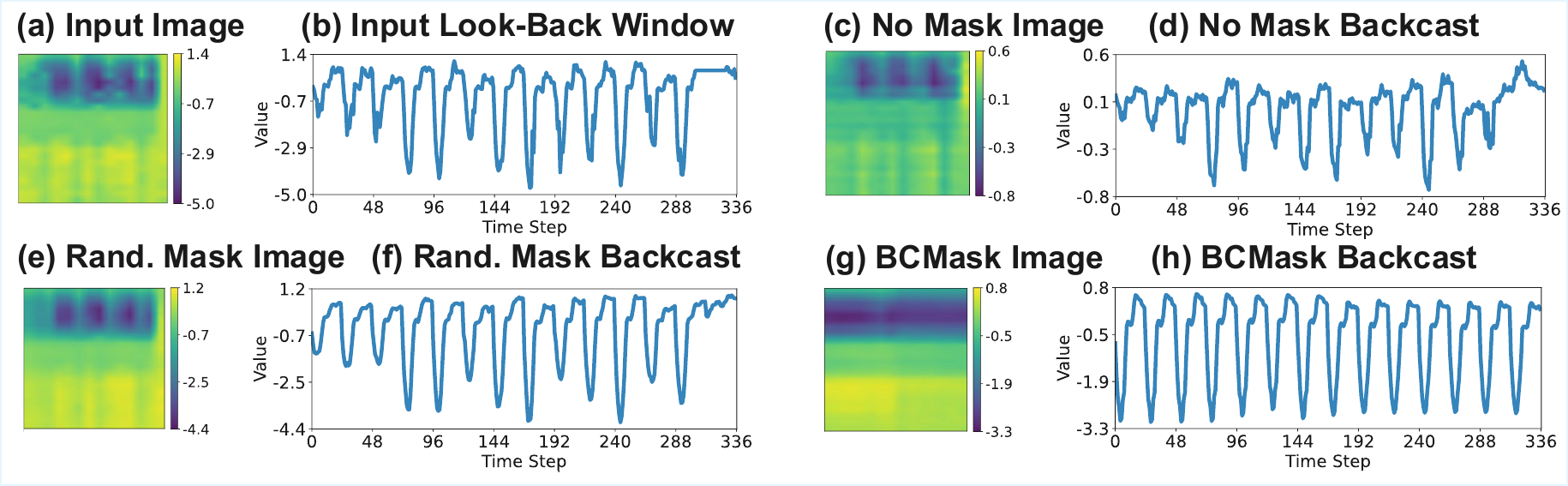}
% \caption{Examples of different masking methods on input series with anomalies.(From ETTh1).}\label{fig.additional_mask_example1}
\caption{Comparison of different masking methods on the same example in ETTh1. (a) image of input look-back window; (c)(e)(g) are images of backcast output by \methodb: (c) uses ``No mask''; (e) uses ``Random mask''; (g) uses \maskmethod. (b)(d)(f)(h) are their recovered time series, respectively.}\label{fig.additional_mask_example1}
\vspace{-0.2cm}
\end{figure*}

\begin{figure*}[!h]
\centering
\includegraphics[width=0.75\linewidth]{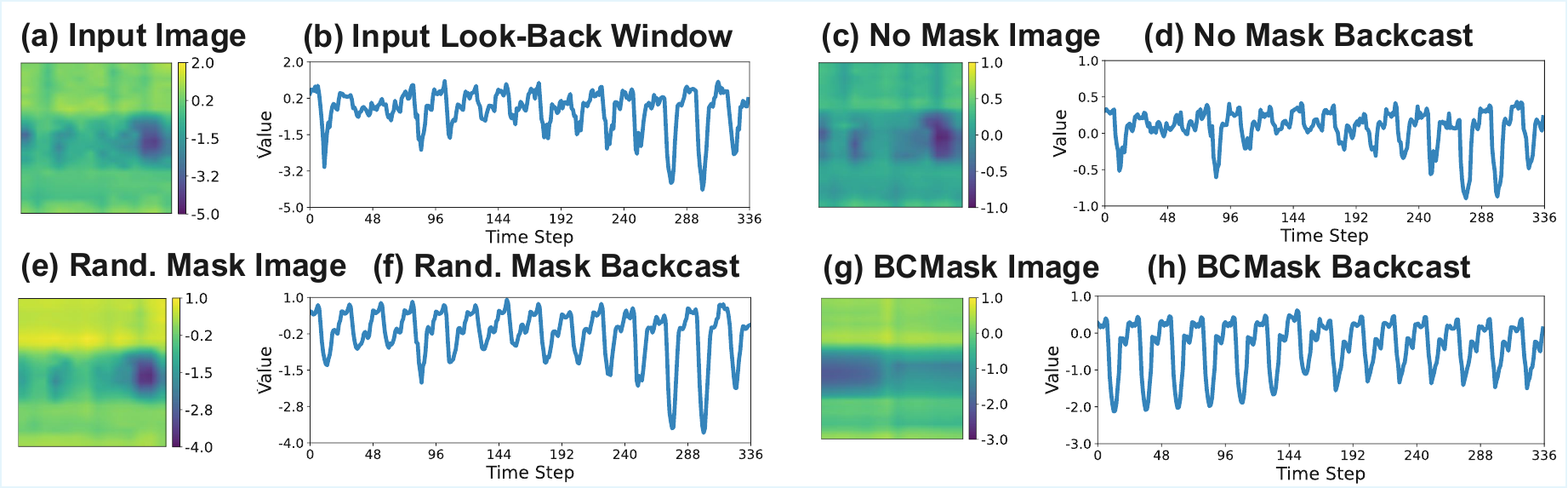}
% \caption{Examples of different masking methods on ETTh2 datasets.}\label{fig.additional_mask_example2}
\caption{Comparison of different masking methods on the same example in ETTh2. (a) image of input look-back window; (c)(e)(g) are images of backcast output by \methodb: (c) uses ``No mask''; (e) uses ``Random mask''; (g) uses \maskmethod. (b)(d)(f)(h) are their recovered time series, respectively.}\label{fig.additional_mask_example2}
\vspace{-0.2cm}
\end{figure*}

Fig.~\ref{fig.additional_mask_example1} and Fig.~\ref{fig.additional_mask_example2} present additional examples of \maskmethod\ in \methodb. Similar to $\S$\ref{sec.exp.analysis}, Fig.~\ref{fig.additional_mask_example1} and Fig.~\ref{fig.additional_mask_example2} compare different masking methods. From both figures, we observe that \maskmethod\ produces smooth patterns along the temporal ($x$-axis) dimension, effectively capturing periodic structures. Notably, when the input time series contains an anomaly ({\em e.g.}, Fig.~\ref{fig.additional_mask_example1}, time steps 288-336), \maskmethod\ can effectively extract the periodic patterns.% Fig.~\ref{fig.additional_mask_example2} provide example from another datasets, demonstrate the generalization ability of \maskmethod.

\subsection{Impact of Look-Back Window}\label{app. Look back windows}

\begin{figure*}[!t]
\centering
\includegraphics[width=0.85\linewidth]{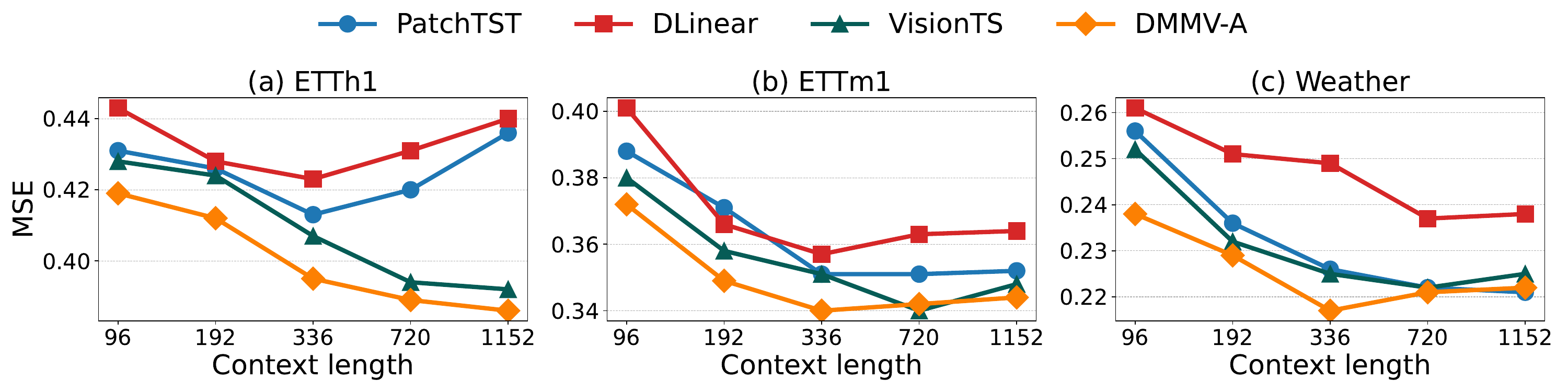}
% \vspace{-1em}
\caption{Average MSE comparison with varying look-back window (or context) lengths.% The MAE is averaged over different prediction lengths.
}\label{fig.context_MSE}
% \vspace{-0.3cm}
\end{figure*}

Fig.~\ref{fig.context_MSE} provides the MSE results that compare \methodb\ with the other three models. Fig.~\ref{fig.context_MSE} demonstrate a similar trend as that of the MAE results in Fig. \ref{fig.context}.% We observe that both \methodb\ and \texttt{VisionTS} continue to benefit from longer look-back windows, while the two numerical-view methods exhibit a significant performance drop beyond a context length of 336. Notably, \methodb\ consistently outperforms all models across different context lengths, highlighting the advantage of our method.

\subsection{Standard Deviations}
\label{app. standard deviations}

To assess the uncertainty and stability of the forecasting performance, we report the standard deviations of \methoda\ and \methodb\ on the four benchmark datasets used in $\S$\ref{sec.exp.ablation} and $\S$\ref{sec.exp.analysis} in Table~\ref{tab. standard deviations}. %This provides a more comprehensive evaluation of the model’s reliability and generalization ability.
From Table~\ref{tab. standard deviations}, the relative standard deviations of the proposed models, which are calculated as the ratio between standard deviation and mean, are all below $1.30\%$ across different datasets and evaluation metrics, demonstrating their stability and robustness over different runs.

\begin{table}[!t]
\centering
\caption{Standard Deviations of \methoda\ and \methodb\ in terms of MSE and MAE on four LTSF benchmark datasets.}\label{tab. standard deviations}
\resizebox{0.6\textwidth}{!}{
\begin{tabular}{c|c|cc|cc}
\toprule[1pt]
\multicolumn{2}{l|}{Model}  & \multicolumn{2}{c|}{DMMV-A}  & \multicolumn{2}{c}{DMMV-S}                                                                      \\\midrule
\multicolumn{2}{l|}{Metric} & MSE & MAE & MSE & MAE \\\midrule
\multirow{2}{*}{\rotatebox{90}{ETTh1}}   & 96     & 0.354$\pm$ 0.001  & 0.390$\pm$ 0.001 & 0.350 $\pm$ 0.001 & 0.388 $\pm$ 0.002 \\
                                         & 192    & 0.393$\pm$ 0.001  & 0.405$\pm$ 0.001 & 0.399 $\pm$ 0.002 & 0.420 $\pm$ 0.001 \\
                                         & 336    & 0.387$\pm$ 0.001  & 0.413$\pm$ 0.001 & 0.401 $\pm$ 0.002 & 0.415 $\pm$ 0.001 \\
                                         & 720    & 0.447$\pm$ 0.002  & 0.451$\pm$ 0.001 & 0.472 $\pm$ 0.001 & 0.480 $\pm$ 0.002 \\ \midrule
\multirow{2}{*}{\rotatebox{90}{ETTm1}}   & 96     & 0.278$\pm$ 0.001  & 0.329$\pm$ 0.000 & 0.296 $\pm$ 0.001 & 0.348 $\pm$ 0.002 \\
                                         & 192    & 0.317$\pm$ 0.001  & 0.358$\pm$ 0.001 & 0.328 $\pm$ 0.001 & 0.368 $\pm$ 0.002 \\
                                         & 336    & 0.351$\pm$ 0.001  & 0.381$\pm$ 0.000 & 0.367 $\pm$ 0.002 & 0.393 $\pm$ 0.002 \\
                                         & 720    & 0.411$\pm$ 0.000  & 0.415$\pm$ 0.000 & 0.401 $\pm$ 0.002 & 0.415 $\pm$ 0.003 \\\midrule
\multirow{2}{*}{\rotatebox{90}{Illness}} & 24     & 1.409$\pm$ 0.001  & 0.754$\pm$ 0.001 & 1.638 $\pm$ 0.003 & 0.842 $\pm$ 0.005 \\
                                         & 36     & 1.291$\pm$ 0.002  & 0.742$\pm$ 0.003 & 1.329 $\pm$ 0.012 & 0.751 $\pm$ 0.002 \\
                                         & 48     & 1.499$\pm$ 0.002  & 0.810$\pm$ 0.011 & 1.643 $\pm$ 0.002 & 0.853 $\pm$ 0.005 \\
                                         & 60     & 1.430$\pm$ 0.003  & 0.774$\pm$ 0.001 & 1.473 $\pm$ 0.002 & 0.810 $\pm$ 0.002 \\\midrule
\multirow{2}{*}{\rotatebox{90}{Weather}} & 96     & 0.143$\pm$ 0.001  & 0.196$\pm$ 0.002 & 0.168 $\pm$ 0.001 & 0.218 $\pm$ 0.002 \\
                                         & 192    & 0.187$\pm$ 0.001  & 0.245$\pm$ 0.003 & 0.221 $\pm$ 0.002 & 0.259 $\pm$ 0.002 \\
                                         & 336    & 0.237$\pm$ 0.001  & 0.272$\pm$ 0.003 & 0.267 $\pm$ 0.002 & 0.305 $\pm$ 0.001 \\
                                         & 720    & 0.300$\pm$ 0.002  & 0.318$\pm$ 0.003 & 0.323 $\pm$ 0.001 & 0.341 $\pm$ 0.003 \\
\bottomrule[1pt]
\end{tabular}
}
\end{table}

\end{document}